\documentclass[sigconf]{acmart}

\AtBeginDocument{%
  \providecommand\BibTeX{{%
    \normalfont B\kern-0.5em{\scshape i\kern-0.25em b}\kern-0.8em\TeX}}}

\setcopyright{acmcopyright}

\copyrightyear{2023}
\acmYear{2023}
\setcopyright{acmlicensed}\acmConference[WWW '23]{Proceedings of the ACM
Web Conference 2023}{April 30-May 4, 2023}{Austin, TX, USA}
\acmBooktitle{Proceedings of the ACM Web Conference 2023 (WWW '23), April
30-May 4, 2023, Austin, TX, USA}
\acmPrice{15.00}
  
\usepackage{microtype}      
\usepackage{xcolor}         
\usepackage{multirow}
\usepackage{graphicx}
\usepackage{amsmath}
\usepackage{enumitem}
\usepackage{wrapfig}
\usepackage{colortbl}

\newcommand{\model}{KGTransformer}
\definecolor{light-gray}{gray}{0.9}

\begin{document}

\title{Structure Pretraining and Prompt Tuning for \\ Knowledge Graph Transfer}



\author{Wen Zhang}
\authornote{Both authors contributed equally to this research.}
\email{zhang.wen@zju.edu.cn}
\author{Yushan Zhu}
\authornotemark[1]
\email{yushanzhu@zju.edu.cn}
\affiliation{%
\institution{Zhejiang University}
\city{Hangzhou}
\country{China}
}

\author{Mingyang Chen}
\email{mingyangchen@zju.edu.cn}
\author{Yuxia Geng}
\email{gengyx@zju.edu.cn}
\affiliation{%
\institution{Zhejiang University}
\city{Hangzhou}
\country{China}
}

\author{Yufeng Huang}
\email{huangyufeng@zju.edu.cn}
\author{Yajing Xu}
\email{22151361@zju.edu.cn}
\affiliation{%
\institution{Zhejiang University}
\city{Hangzhou}
\country{China}
}

\author{Wenting Song}
\email{songwenting@huawei.com}
\affiliation{%
\institution{Huawei Technologies Co., Ltd}
\city{Xi'an}
\country{China}
}

\author{Huajun Chen}
\authornote{Corresponding author.}
\email{huajunsir@zju.edu.cn}
\affiliation{
\institution{Zhejiang University}
\institution{Donghai laboratory}
\institution{Alibaba-Zhejiang University Joint Institute of Frontier Technology}
\city{ }
\country{ }
}

\renewcommand{\shortauthors}{ Zhang and Zhu, et al.}

\begin{abstract}
  Knowledge graphs (KG) are essential background knowledge providers in many tasks. When designing models for KG-related tasks, one of the key tasks is to devise the Knowledge Representation and Fusion (KRF) module that learns the representation of elements from KGs and fuses them with task representations. While due to the difference of KGs and perspectives to be considered during fusion across tasks, duplicate and ad hoc KRF modules design are conducted among tasks. In this paper, we propose a novel knowledge graph pretraining model KGTransformer that could serve as a uniform KRF module in diverse KG-related tasks. We pretrain KGTransformer with three self-supervised tasks with sampled sub-graphs as input. For utilization, we propose a general prompt-tuning mechanism regarding task data as a triple prompt to allow flexible interactions between task KGs and task data. We evaluate pretrained KGTransformer on three tasks, triple classification, zero-shot image classification, and question answering. KGTransformer consistently achieves better results than specifically designed task models. Through experiments, we justify that the pretrained KGTransformer could be used off the shelf as a general and effective KRF module across KG-related tasks. The code and datasets are available at https://github.com/zjukg/KGTransformer.
\end{abstract}



\maketitle

\section{Introduction}
Knowledge Graphs (KG) representing facts as triples in the form of \textit{(head entity, relation, tail entity)}, abbreviated as \textit{(h,r,t)}, is a common way of storing knowledge in the world, such as \textit{(Earth, location, inner Solar System)}\footnote{Example from Wididata.}.
In recent years, many large-scale KGs including Wididata \cite{wikidata}, YAGO \cite{yago} and NELL \cite{nell} have been constructed and applied as background knowledge providers in machine learning tasks, such as question answering \cite{qagnn}, image classification \cite{gcnz}, visual reasoning \cite{DBLP:conf/emnlp/GarderesZAL20}, etc. 

\begin{figure*}[t]
    \centering
    \includegraphics[width = 0.85\linewidth]{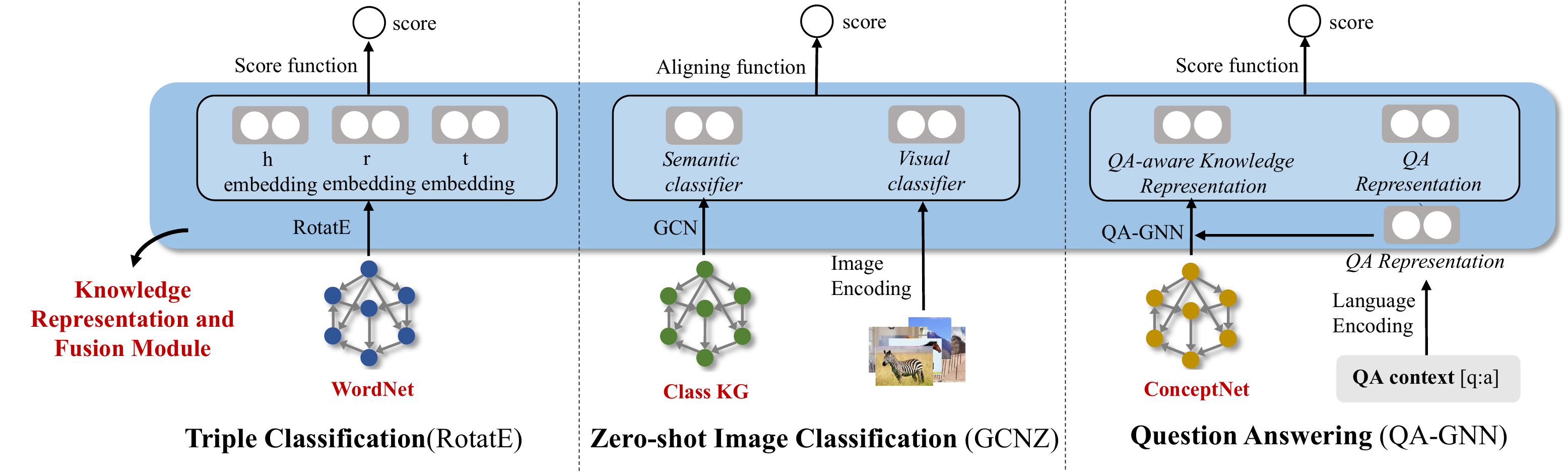}
    \vspace{-3mm}
    \caption{Example of models for KG-related tasks supported by different KGs.}
    \vspace{-3mm}
    \label{fig:motivation}
\end{figure*}

When designing models for KG-related tasks, 
one of the key tasks is to devise the Knowledge Representation and Fusion (KRF) module that learns representation of elements from KGs and fuses them with task representations. 
As shown in Figure~\ref{fig:motivation},
the knowledge graph completion model RotatE \cite{rotate} 
represents knowledge in KGs by learning embedding for entities and relations in the complex value space and calculates the truth value of triples through a score function.
The zero-shot image classification model GCNZ~\cite{gcnz} uses a graph convolutional network and ResNet \cite{resnet} to learn representations of KGs and images, respectively, 
and fuses them through aligning results from classifiers on KG representation and image representation.
The knowledge-based question answering model QA-GNN \cite{qagnn} first encodes the query-KG graph and fuses the representation with query representation for prediction.
\textbf{Due to the difference of KGs and perspectives to be considered, the KRF modules are different in KG-related task models, leaving duplicate works for ad hoc KRF module design.}

To solve this problem,
the pre-trained KG model is proposed to learn universal embeddings of entities and relations that could be applied in many tasks \cite{conceptnet,pkgm}. These embeddings are supposed to contain entity similarities \cite{transe}, hierarchies \cite{hake} and relationships \cite{complex} that could be used for recommender system \cite{KGAT}, entity alignment \cite{DBLP:conf/ijcai/SunHZQ18}, question answering \cite{QA-KG}, etc., helping implicitly access knowledge in KGs. However, directly applying  embeddings from a pre-trained model is insufficient and meets two challenges:
(1) The first challenge is 
if task KGs contain different entities and relations to KGs used for pre-training, 
the embedding-based method could not transfer valuable information to downstream task model since the embeddings are missing. 
(2) The second challenge is essential interaction and fusion between KGs and task data is missing, leaving designing a fusion module as part of the work for downstream task model devising. 
\textbf{Thus embedding-based KG models are not ideal solutions for KRF module across KG-related tasks}.

In this work, we solve the first challenge by pre-training KG structures and transforming parameters unrelated to specific entities and relations into tasks. 
We solve the second challenge by a prompt tuning mechanism to enable uniform and flexible fusion between KGs and task data. 
As a result, \textbf{we propose a novel model {\model},
constructed by multiple \textit{ {\model} layers} with a sequence of triples as input}. It allows diverse but constrained interactions between elements in the sequence according to a neighborship matrix between elements. 
We propose a sub-graph pre-training method with three self-supervised tasks, i.e. Masked Entity Modeling, Masked Relation Modeling, and Entity Pair Modeling.
This enables {\model} to capture graph structures and semantics universally existing in KGs. 
The set of parameters $\theta_{\mathcal{M}}$ in {\model} layers helps transfer graph structural knowledge that is unrelated to specific entities and relations from pre-training KGs to tasks KGs. 
For applying {\model} in KG-related tasks,
we propose a general prompt tuning mechanism forming each task sample as a prompt concatenated at the end of the task KG sequence to manipulate the performance of {\model}.  

During experiments, we pre-train \model~on a hybrid dataset consisting of three benchmark datasets with diverse structures. Then we apply it to three KG-related tasks of different modalities, including one in-KG task triple classification, and two out-of-KG tasks, zero-shot image classification and question answering. We compare {\model} to recently proposed and specifically designed methods of these three tasks. Results show that {\model} performs better, proving the effectiveness of the pre-trained {\model} as a uniform KRF module for KG-related task models. 
More importantly, we prove that simply using pre-trained {\model} layers off the shelf with $\theta_{\mathcal{M}}$ frozen in tasks is enough to get promising results. In summary, our contributions are 
\begin{itemize}
    \item We propose the novel {\model}, 
    which could capture graph structural knowledge that is transferable across KGs by being pre-trained with self-supervised tasks. 
\item We propose a simple yet effective prompt tuning mechanism to apply  {\model} off the shelf to enable flexibly fusing knowledge in KGs to task data. 
\item We show that 
the pre-trained {\model} has the capability of transferring KG structure knowledge across KGs and is general enough to be applied in various tasks, supported by experiments on three KG-related tasks.
\end{itemize}

\section{Related Works}
\subsection{Knowledge Graph Representation Methods}
KG representation methods encode information in KGs through parameters and functions in models. They could recover the graph structures and capture semantics between entities and relations.

\textit{Embedding-based methods} \cite{transe,transr,complex,rotate} learn embeddings of relations and entities and model the truth value of triples through a score function with  embeddings as inputs. After training, these embeddings could implicitly capture the similarities \cite{transe}, hierarchies \cite{hake}, relationships \cite{rotate}, and axioms \cite{itere} between elements in KGs, thus could be applied as general representations of elements in many tasks to transfer semantics learned from KGs to tasks. 

\textit{Structure-based methods} \cite{grail, compile,inductive-ijcai,inductive-sigir} learn an encoder with sub-graph with node features as input, and a decoder for specific tasks. Typical graph-based methods are independent of embeddings of entities and thus are entity agnostic that could be applied to inductive tasks. For example, GraIL \cite{grail} generates two-dimensional node features for each entity in a sub-graph, encodes the sub-graph through a graph-neural-network-liked module to get the graph-specific representation of entities, and uses the entity and relation representations from encoder for link prediction. 

\textit{Hybrid-based methods} \cite{rgcn,compgcn,hran,M-GNN} learn encoder-decoder together with embeddings of entities and relations. Specifically, embeddings of entities rather than pre-defined node features are used for graph neural network (GNN) encoder. Our {\model} is also a hybrid-based method regarding KG representation. While different from existing hybrid-based methods, apart from one-hop neighbors of entities, {\model} also aggregates information from connected relations and two-hop neighbors.

\textit{Transformer-based methods} \cite{DBLP:conf/sigir/ChenZLDTXHSC22,DBLP:conf/emnlp/ChenLG0ZJ21,DBLP:conf/kdd/LiuZSCQZ0DT22,DBLP:conf/emnlp/SunGZ0022,DBLP:conf/naacl/Koncel-Kedziorski19,DBLP:conf/kdd/LiZZ22,DBLP:conf/emnlp/HuGXLP22} encode structural or semantic information in KGs through attention mechanisms \cite{transformer}. GraphWriter \cite{DBLP:conf/naacl/Koncel-Kedziorski19} proposes a graph Transformer to solve the problems of non-hierarchical nature, long-distance dependencies, and structural variation for generating text from KGs. HittER \cite{DBLP:conf/emnlp/ChenLG0ZJ21} proposes a hierarchical Transformer to jointly learn entity-relation composition and relation contexts in KGs. kgTransformer \cite{DBLP:conf/kdd/LiuZSCQZ0DT22} is pretrained on KGs by formulating logical queries as masked prediction and applied to complex query tasks. GHT \cite{DBLP:conf/emnlp/SunGZ0022} is proposed for temporal KGs reasoning and it utilizes Transformers to capture the instantaneous structure and evolution information. MKGformer \cite{DBLP:conf/sigir/ChenZLDTXHSC22} designs a hybrid Transformer that integrates visual and text representation for multimodal KGs completion tasks. TET \cite{DBLP:conf/emnlp/HuGXLP22} performs the entity typing task by encoding KGs' graph structure through a local Transformer, a global Transformer, and a context Transformer. 

\subsection{Knowledge Graph Fusion Methods}

Many knowledge graphs are proposed in various domains and are used to support different downstream tasks. As a kind of important side information, researchers design sophisticated ways to incorporate knowledge graphs into their task-specific methods, and we term them knowledge graph fusion methods.

Some works \cite{ontozsl,embedKGQA,BiNet,DOZSL,ENeSy,StAR} use knowledge in KGs in an \textit{out-of-the-box} manner. Specifically, these methods usually conduct representation learning on KGs in advance and use trained representations of entities and relations as the input of downstream models or for ensembling; such pretrained embedding can be frozen or fine-tuned during training downstream models.
For example, OntoZSL \cite{ontozsl} uses trained TransE \cite{transe} embeddings for an ontological schema to model the prior knowledge for zero-shot learning. EmbedKGQA\cite{embedKGQA} uses trained ComplEx \cite{complex} embeddings to support the answer selection in question-answering systems.

Furthermore, other works \cite{RippleNet,KGAT,qagnn,KGT5,kepler} fuse knowledge in an \textit{end-to-end} manner. More precisely, these methods design learnable KG encoder modules and train them with downstream models. 
For instance, RippleNet \cite{RippleNet} encodes the sub-graph preference propagation in a KG and predicts the user engaging in recommender systems. 
KGAT \cite{KGAT} employs an attention mechanism to propagate representations from neighbors for nodes in a KG containing users, items, and attributes for the recommendation.
QA-GNN \cite{qagnn} uses a GNN to encode QA-pair-related sub-graphs for joint representation of QA text information and KG information. Our {\model} is also an end-to-end KG fusion method. 

\subsection{Pretraining Methods}
The great success of pretrained language models (PLMs) \cite{bert,roberta,bart,radford2018improving} has shifted the paradigm of natural language processing from \textit{fully supervised learning} to \textit{pretraining and fine-tuning} \cite{liu2021pre}. PLMs with fixed architectures are pretrained on large-scale corpora to learn robust general-purpose features of a language. By adding additional parameters, PLMs could be quickly adapted to downstream tasks optimized according to the task-specific objective function. The paradigm of pretraining and fine-tuning is widely adopted in other areas, such as image processing \cite{beit}, multimodal processing \cite{mm-pretrain1,mm-pretrain2}, and table processing \cite{table1, table2}. Since general knowledge such as axioms also exists in knowledge graphs, this inspires us to explore pretrained knowledge graph models so as to be quickly adapted to KG-related tasks. 

Among pretraining methods, 
the auto-regressive \textit{Transformer} \cite{transformer} with multi-head self-attention module is the key part of pretrained language models \cite{bert,roberta}. It has been shown to be very powerful in processing sequential data such as text. 
Even for non-sequential data such as image \cite{vilbert}, video \cite{vivit} and graph \cite{graphomer},  
Transformer with specific designs, such as  task data serialization \cite{kgbert}, position encoding \cite{vit}, structure encoding \cite{graphomer}, etc., could also perform reasonably well. 
To adapt Transformer to graph data, Ying et al. \cite{graphomer} propose Graphomer built on the standard Transformer with several structural encoding methods, including spatial encoding, edge encoding, and centrality encoding. 
In the works adapting Transformer to encode KGs, concatenating surface form or description of the head entity, relation, and tail entity to serialize triple are commonly applied \cite{kgbert,DBLP:conf/coling/KimHKS20}, while they pay more attention to the text of elements than graph structures. 
In contrast to encoding graph structure as it is or allowing connections between all elements, the {\model} layer we propose in this paper allows constrained connections between parts of elements that are not directly connected in KG.

During finetuning, the \textit{prompt tuning} methods on PLM introduce a flexible and effective way to manipulate the behavior of PLMs \cite{prompttuning,DBLP:conf/acl/GaoFC20,DBLP:journals/jmlr/RaffelSRLNMZLL20} by adding an appropriate prompt related to the tasks. 
They inspire us to adapt pretrained knowledge graph models by adding the task sample as a prompt. 

\section{Methodology}

In this section, we introduce the model structure of {\model} (\ref{sec:kgtransformer}), how to pretrain {\model} with graph sampling strategies and pretraining tasks (\ref{sec:sub-graph-pre-training}), and the prompt-tuning mechanism for KG-related tasks (\ref{sec:task-prompt-tuning}). 

\begin{figure*}[t]
    \centering
    \includegraphics[width = 0.9 \linewidth]{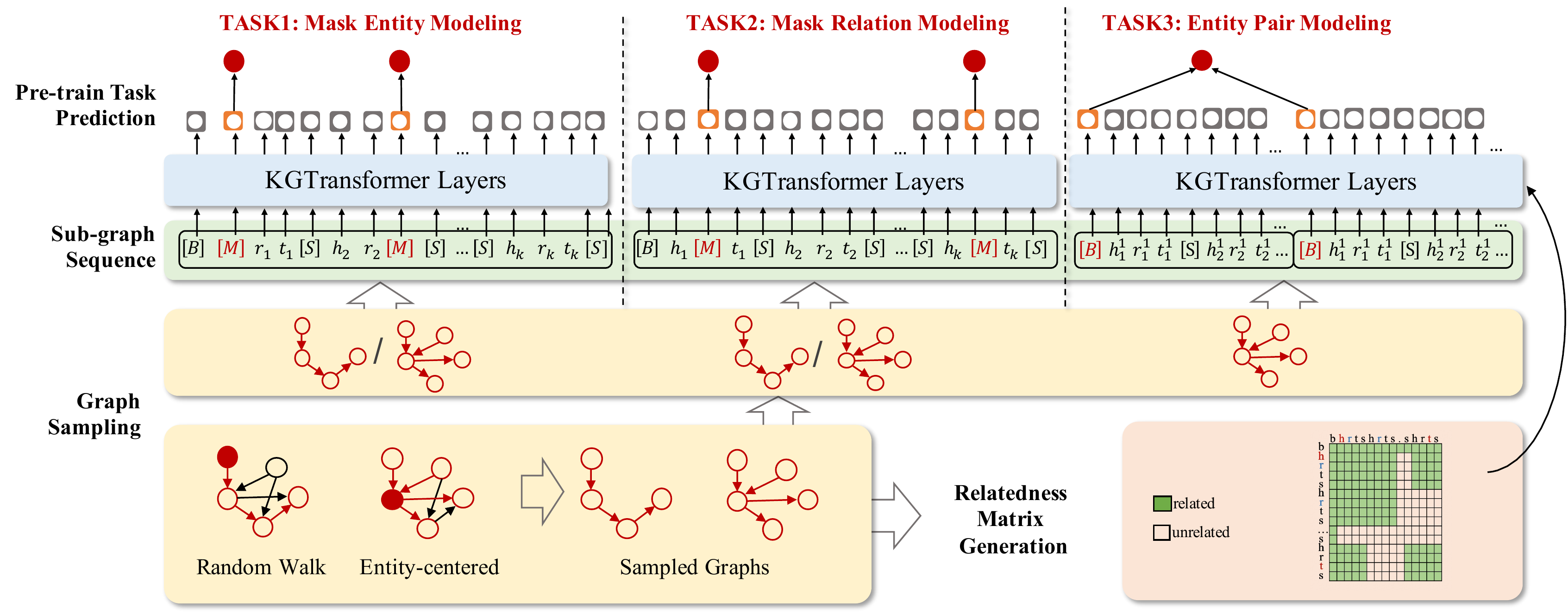}
    \caption{Overview of sub-graph pretraining of {\model}.}
    \label{fig:pre-train}
\end{figure*}

\subsection{KGTransformer}
\label{sec:kgtransformer}
{\model} is constructed by multiple {\model} layer built on traditional Transformer layer \cite{transformer}. We adapt Transformer layer to  knowledge graph with a set of triples $\mathcal{T}_{in} = \{(h_i, r_i, t_i) | i \in [1, k]\}$ as input and output the representation of each element that could be used for prediction, where $k$ is the number of triples. 

Given a $\mathcal{T}_{in}$,
we first make $\mathcal{T}_{in}$ as a sequence of tokens
\begin{equation}
\label{equ:sequence}
    s_{in}= [ [B],h_1, r_1, t_1,[S], h_2, r_2, t_2, [S], ...,h_k, r_k, t_k, [S] ] 
\end{equation}
where $[B]$ and $[S]$ are special tokens 
indicating the beginning of sequence and separation between triples respectively. 

Then we generate a matrix $M \in \mathbb{R}^{|s_{in}| \times |s_{in}|}$ to indicate the neighborships between triples that are not explicitly modeled in the sequence $s_{in}$, 
where $|s_{in}|$ is the length of $s_{in}$. 
\begin{align}
    & M_{ij} = \left\{ 
    \begin{aligned} 
    & 1 \text{ if } i=1 \text{ or } j=1 \\
    & 1 \text{ if }  trp(i)\cap trp(j) \ne \emptyset \\ 
    & 0 \text{ otherwise } \\
    \end{aligned} 
    \right., 
    \; \\
    & trp(n) = \{h_{p}, r_p, t_p\} \text{ where } p = \lfloor(n-2)/4\rfloor+1 
    \label{matrix}
\end{align}
With $s_{in}$ and $M$,
we input $s_{in}$ to the {\model } and use $M$ to constrain the interactions between elements in $s_{in}$. Specifically, similar to traditional transformer, {\model } layer includes a self-attention module and a position-wise feed-forward network. 
Suppose the input of self-attention module is $H=[ s_1^{\top}, ..., s_n^{\top}]^\top \in \mathbb{R}^{n\times d}$
with the $i$th row as the $d$ dimensional hidden state for the $i$th element in the sequence. The self-attention operation $Attn()$ is 
\begin{align}
    & Q = HW_{Q}, \; K = HW_{K}, \; V=HW_{V}, \\
    & A = \frac{QK^{\top}\odot M}{\sqrt{d_K}} + (1-M)*\delta, \\
    & Attn(H) = softmax(A)V,
    \label{equ:transformer}
\end{align}
where $W_{Q}\in \mathbb{R}^{d\times d_Q}$,$W_{K}\in \mathbb{R}^{d\times d_K}$,$W_{V}\in \mathbb{R}^{d\times d_V}$ is the projection matrix to generate the query, key, and value representation of $H$; 
$\odot$ represents element-wise multiplication;
$A$ is the attention matrix with $A_{ij}$ capturing the similarity between the query representation of $s_i$ and key representation of $s_j$;
$\delta$ is a large negative number to make the $A_{ij}$ with $M_{ij}=0$ near to $0$ after softmax function. 
Following traditional transformer, multi-head self-attention is applied in each {\model} layer. And the input of the first {\model } layer is sequential embedding of elements in $s_{in}$.

Here we discuss benefits of the {\model} layer with $s_{in}$ as input sequence.
(1) Compared to conventional Transformer which allows each element to attend to all elements in the sequence, {\model} avoids attention between elements in unrelated triples such as \textit{(William Shakespeare, field of work, Fiction)} and \textit{(France, capital, Paris)} since they do not share any element. 
(2) Compared to conventional graph neural networks \cite{gat,rgcn} that explicitly encode the graph structures via aggregate one-hop neighbors to entities,  
{\model} allows aggregation from  one-hop and two-hop neighbors to each entity per update. For example, considering triple \textit{(William Shakespeare, field of work, Fiction)} and \textit{(William Shakespeare, notable book, Hamlet)},
\textit{Hamlet} attends to one-hop neighbor \textit{William Shakespeare} and two-hop neighbor \textit{Fiction} in {\model} layer while only attends to \textit{William Shakespeare} in traditional graph neural network.
(3) {\model} layer enables entities to attend to elements in triples that are not directly connected but share the same relation, for example,  \textit{(China, capital, Beijing)} and \textit{(France, capital, Paris)}.
In summary, {\model} layer allows diverse but constrained element interactions in KGs. 

Finally, {\model} is constructed by stacking $m$ {\model} layers. 

\begin{figure*}[t]
    \centering
    \includegraphics[width = 0.9\linewidth]{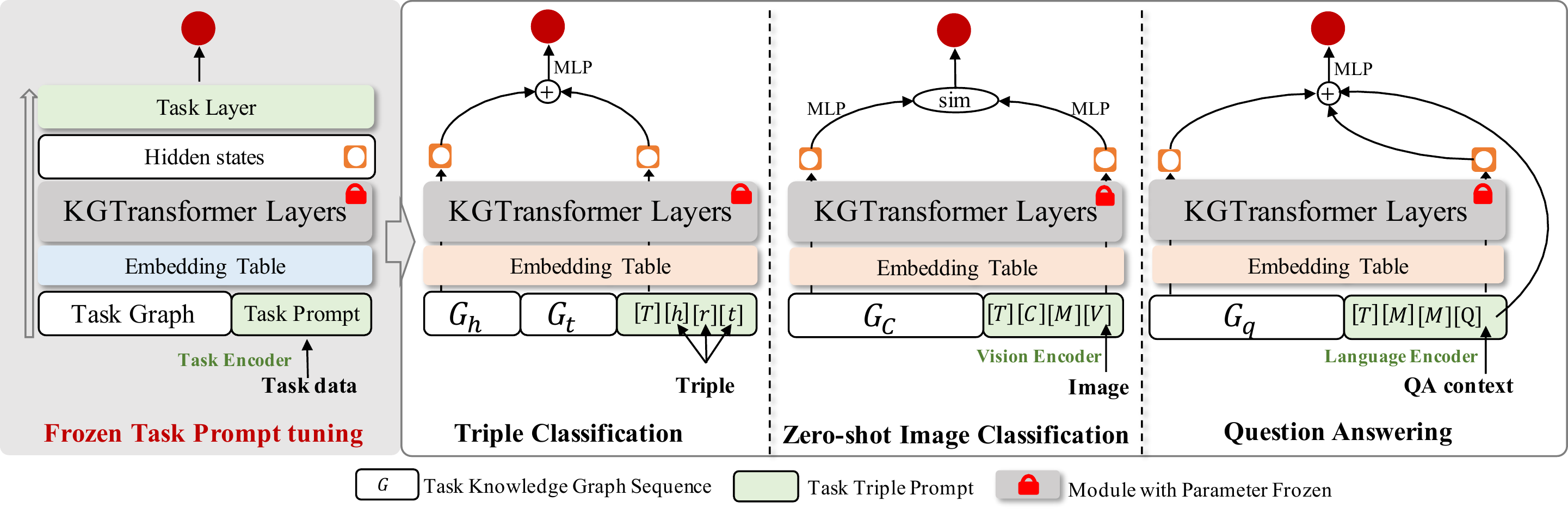}
    \vspace{-3mm}
    \caption{Overview of task prompt tuning (left) and examples of three specific tasks (right).}
    \vspace{-3mm}
    \label{fig:fine-tune}
\end{figure*}

\subsection{Sub-graph Pretraining}
\label{sec:sub-graph-pre-training}
Given a knowledge graph $\mathcal{G}$, 
we propose to pretrain the {\model} 
by self-supervised tasks with sampled sub-graphs from $\mathcal{G}$.
The overall procedure is shown in Figure \ref{fig:pre-train}. 

\subsubsection{Sub-graph Sampling}
We propose two strategies to sample a sub-graph $\mathcal{G}_e$ starting from a randomly selected $e$ as target entity.

\begin{itemize}[leftmargin=*]
    \item \textit{Random Walk Sampling}. We randomly choose a triple $(e, r, e^\prime) \in \mathcal{G}$ or $(e^\prime, r, e) \in \mathcal{G}$, and add it into  $\mathcal{G}_{e}$. Then regarding $e^\prime$ as 
    target entity and repeat this step for $k$ times. 
    Such sampling captures global and long sequential relatedness between triples in KGs. 
    \item \textit{Entity-centered Sampling}. We randomly choose $max(k,n)$ triples from one-hop triple set $\{(e, r, e^\prime) \in \mathcal{G}\} \cup \{ (e^\prime, r, e) \in \mathcal{G}\}$, where $n$ is the number of one-hop triples of $e$.
    If one-hop triples are not sufficient that $n<k$, 
    we sample $k-n$ two-hop triples from $\{(e^{\prime}, p, e^{\prime\prime}) \in \mathcal{G}\} \cup \{ (e^{\prime\prime}, p, e^{\prime}) \in \mathcal{G}\}$. Entity-centered sampling could capture local relatedness between triples. 
\end{itemize}
After sampling, we serialize the sub-graph $\mathcal{G}_{e}$ into a sequence $s_{in}$ following Equation (\ref{equ:sequence}) as the input of {\model}. 

\subsubsection{Pretraining Tasks.} 
In order to make {\model} capture the graph structures and semantics of elements in KGs, we propose three self-supervised tasks. In the rest of the paper, we use $s_e^m$ to represent the hidden state of element $e$ from the last layer of {\model}.
\begin{itemize} [leftmargin=*]
    \item \textit{Masked Entity Modeling (MEM)}.  
    We randomly replace some entities $ \mathcal{M}_e$ with mask token $[M]$. For each masked entity $e$, we make projected $s_e^m$ via $W_{EME} \in \mathbb{R}^{d\times d}$ to be closed to $\mathbf{E}(e)$, the embedding of entity $e$ before being input to the first KGTransformer layer, and far away from others. The loss of MEM is
    \begin{align}
        L_{MEM}(\mathcal{G}_e) = & \sum_{e \in {\mathcal{M}_e}}CE(s_e^m W_{MEM}\mathbf{E}(e)^\top,1) \\ 
        & + \sum_{e^\prime \in \bigtriangleup}CE(s_e^m W_{MEM}\mathbf{E}(e^\prime)^\top,0),
        \label{equ_MEM}
    \end{align}
    where CE() is the cross entropy loss function. $\bigtriangleup = \{e' | e'\ne e \}$ is the set of negative entities that are not equal to masked entity $e$.
    \item \textit{Masked Relation Modeling (MRM)}.
    We randomly replace some relations $\mathcal{M}_r$ with $[M]$. For each masked relation $r$, we conduct multi-classes classification and the loss is 
    \begin{equation}
        L_{MRM}(\mathcal{G}_e) = \sum_{r\in \mathcal{M}_r} CE(MLP(s_r^mW_{MRM}), l_{r}), \; W_{MRM}\in \mathbb{R}^{d \times d}
        \label{equ_MRM}
    \end{equation}
    where 
    $W_{MRM} \in \mathbb{R}^{d\times n_r}$ is a transformation matrix; 
    $l_{r}$ is a one-hot vector label for masked relation $r$. 
    \item \textit{Entity Pair Modeling (EPM)}.  
    Given two sub-graphs $\mathcal{G}_{e_i}$ and ${\mathcal{G}_{e_j}}$ of $e_i$ and $e_j$, we serialize and concatenate them as input sequence and the loss function is 
    \begin{equation}
        L_{EPM}(\mathcal{G}_{e_i}, \mathcal{G}_{e_j}) = CE(MLP([s_{[B]_{e_i}}^m||s_{[B]_{e_j}}^m ]),l_{(e_i, e_j)})
        \label{equ_EPM}
    \end{equation}
    where $s_{[B]_{e_i}}^m$ is the hidden state from the last {\model} layer corresponding to $[B]$ in the sequence created from $\mathcal{G}_{e_i}$. $[x||y]$ represents the concatenation of vector $x$ and vector $y$. $l(e_i, e_j)$ is the label for similarity of $e_i$ and $e_j$. We regard entity $e_i$ and entity $e_j$ to be similar and labeled them $l(e_i, e_j) =1$ if they used to be the head(tail) entity of the same relation in $\mathcal{G}$, otherwise $l(e_i, e_j) =0$. 
\end{itemize}
With these three self-supervised tasks, 
the overall pretraining loss function is
    \begin{equation}
    L(\mathcal{G})=\sum_{e \in \mathcal{E}} (L_{MEM}(\mathcal{G}_e^1)+L_{MRM}(\mathcal{G}_e^2)+L_{EPM}(\mathcal{G}_e^3, \mathcal{G}_{e'}))
        \label{equ_pretrain}
    \end{equation}
where $\mathcal{G}^1_e$ and $\mathcal{G}_e^2$ are subgraphs of $e$ from random walk sampling or entity-centered sampling; $\mathcal{G}_e^3$ and $\mathcal{G}_{e'}$ is subgraphs of $e$ and $e'$ from entity-centered sampling, where $e' \ne e$. 
\subsection{Task Prompt Tuning}
\label{sec:task-prompt-tuning}
With pretrained {\model}, we propose a simple and uniform prompt tuning mechanism to apply it in KG-related tasks. 

Before prompt tuning, we continually train {\model} with KGs in tasks following the pretraining method introduced in Section \ref{sec:sub-graph-pre-training}, with $\theta_{\mathcal{M}}$ frozen, since the entities and relations are different in tasks. After such training, we get the element embeddings of task KGs adapted to the  $\theta_{\mathcal{M}}$. 

During prompt tuning, in a task, one sample includes task data $\mathcal{D}_{task}$ and a supporting graph $\mathcal{G}_{task}$. For example, in question answering, $\mathcal{D}_{task}$ is a question-choice pair and  $\mathcal{G}_{task}$ is a knowledge graph extracted based on keywords in the pair. 
Firstly, with $\mathcal{D}_{task}$, we construct a prompt in the form of a triple sequence that $\mathcal{P}_{task} = [[T] [H] [R] [D]]$
where $[T]$ is a special token indicating the beginning of task which is randomly initialized during tuning. $[H]$ and $[R]$ indicate the head entity and relation in the task triple. $[D]$ is the tail entity in task triple and its representation is from a task encoder with $\mathcal{D}_{task}$ as input. 
Secondly, 
for $\mathcal{G}_{task}$, we serialize it into a sequence in the form as Equation (\ref{equ:sequence}) and concatenate it with task  prompt as the input of {\model}. 
Finally, we take out some hidden state $s^m$ from the last {\model} layer and input them into a task layer for prediction. 
We make parameters in {\model} layers $\theta_{\mathcal{M}}$ frozen to keep the knowledge of graph structure unrelated to specific entities and relations learned during pretraining as it is. 
The overall process of task prompt tuning is shown in left part in Figure \ref{fig:fine-tune}. 
The right part of Figure~\ref{fig:fine-tune} shows examples of three specific tasks. 
Since the fine-tuning processes in different tasks are slightly different, we'll introduce the details of specific tasks in Section~\ref{experiment}.

\begin{table}[]
    \centering
    \caption{Statistics of WFC dataset. Std$_{rel}$(Std$_{ent}$)  is the standard deviation of the number of triples of relations(entities).}
    \vspace{-3mm}
    \resizebox{0.46\textwidth}{!}{ 
    \begin{tabular}{l|c c c c c c }
    \toprule
         & \#R & \#E & \# T &Std$_{rel}$ & Std$_{ent}$ & Density($\times 10^{-6}$)  \\
         \midrule
        \textbf{WFC} & 317 & 133435 & 1015556 & 13230 & 134 & 0.18  \\
        \midrule
       $wn18rr$  & 11 & 40943 & 93003 & 12540 & 9 & 5.04 \\
       $fb15k-237$ & 237 & 14541 & 310115 &  2467 &128 & 6.19 \\
       $codex$ & 69 & 77951 & 612437 & 24664 & 159 & 1.46 \\
       \bottomrule
    \end{tabular}
    }
    \label{tab:dataset-pretrain}
\end{table} 
\begin{table*}[]
    \centering
    \caption{Statistics of dataset in three tasks. \#E, \#R, and \#T is the number of entities, relations, and triples.}
    \vspace{-3mm}
    \resizebox{\textwidth}{!}{
    \begin{tabular}{l|c |c c c |c |c c c | c c c }
        \toprule
        & \multicolumn{4}{c|}{\textbf{Task KG}} & \multicolumn{4}{c|}{\textbf{Task Data}} &\multicolumn{3}{c}{\textbf{Properties}} \\
         \cline{2-12}
         & KG & \#E &\#R &\#T & Task Sample  &\# Train & \#Valid & \#Test & Task Type & Modality & Overlap to WFC \\
        \midrule
        T1: Triple Classification & WN18RR & 40943 & 11 & - &Triple &  86835 & 6068 & 6268 & in-KG & KG & Yes \\
        T2: ZSL Image Classification & AwA-KG & 146 & 16 & 1595 &Image & 23527 & -  &  13795 & out-of-KG & Image+KG & No \\
        T3: Question Answering & CommonsenQA & 64388 & 16 & 309444 & QA pair & 8500 & 1221 & 1241 &out-of-KG  &Language+KG & No \\
        \bottomrule
    \end{tabular}
    }
    \label{tab:dataset}
    \vspace{-3mm}
\end{table*} 

\section{Experiment}
\label{experiment}
In this section, we introduce the experiments of {\model } on KG-related tasks. We choose the tasks based on the principle to cover both in-KG and out-of-KG tasks, data of different modalities, and different overlaps with pretraining dataset. Finally, three tasks, triple classification, zero-shot image classification, and question answering, are selected. 

After pretraining, in each task, we show the results of specifically designed tasks models and {\model } with three settings during task tuning:
re-using $\theta_{\mathcal{M}}$ and keeping them frozen (\textit{{\model}});
tuning $\theta_{\mathcal{M}}$ together with task parameters (\textit{{KGT}-finetune});
training {\model} from scratch (\textit{{KGT}-scratch}). 

\subsection{pretraining}
\subsubsection{Datasets.} 
For pretraining of {\model}, 
a KG covers diverse graph structures is preferred since {\model} is designed for graph structure pretraining. Thus we created a new dataset named WFC by combining three common KG datasets with different graph properties, including WN18RR, FB15k-237, and Codex. WN18RR is a dense and unbalanced KG. FB15k-237 is a dense and balanced KG. Codex is a sparse and unbalanced KG. Containing these three datasets, WFC covers diverse graph structures. The statistics of WFC and three datasets are shown in Table~\ref{tab:dataset-pretrain}.

\subsubsection{Pretraining Details.} 
The {\model } is constructed with 4 {\model} layers. The token embeddings and $\theta_{\mathcal{M}}$ are randomly initialized. 
In each layer, there are $768$ hidden units and $12$ attention heads. The number of triples $k$ in sampled subgraph is set to $126$. Due to the disorder of the triples in the sub-graphs, the position embedding is not applied in {\model}.

During pretraining, given an entity $e$, we generate multiple samples for each task. For MEM and MRM tasks, we get a sub-graph by random-walk sampling or entity-centered sampling and randomly select $15\%$ of triples to mask either the head or the tail entity in the MEM task and to mask relation in the MRM task, when serializing the graph into an input sequence. We randomly sampled $2$ entities in current training batch as $e^\prime$ in Equation (\ref{equ_MEM}) for MEM task.
For the EPM task, we first sample a positive(negative) entity pair ($e$, $e_j$) with a probability of $0.5$, where $e_j$ and $e$ used(not used) to be the head or tail entity of the same relation. We sample the sub-graph of  $e$ and $e_j$ by entity-centered sampling. The MLPs in Equation (\ref{equ_MRM}) and Equation (\ref{equ_EPM}) are both 1-layer.
The model is implemented with Pytorch and trained on 1 Tsela-A100 GPU with batch size as $4$ for $10$ epochs ($97$ hours).
The optimizer is Adam whose learning rate is $0.0001$, together with a linear decay learning rate schedule with warm-up.
The model size is $508$M containing about $133$ million parameters, among which there are  28 million parameters in $\theta_{\mathcal{M}}$, that will transfer to KG-related tasks.

\subsection{Task1: Triple Classification}
\subsubsection{{\model} Implementation.}  
In this task, $\mathcal{D}_{task} = (h,r,t)$ and $\mathcal{G}_{task}$ is the concatenation of $h$ and $t$ centered sub-graph, denoted as $\mathcal{G}_h$ and $\mathcal{G}_t$ respectively. $\mathcal{P}_{task} = [[T]\; [h]\; [r]\; [t] ]$. In the last layer of {\model}, we take out the hidden states corresponding to the first token $[B]$ and task token $[T]$, denoted as $s_{[B]}^m$ and $s_{[T]}^m$ which are inputted into a feed-forward network to output the score of $(h,r,t)$. The score of $(h,r,t)$ is expected to be $1$ for the positive one and $0$ for the negative one. We use a cross-entropy loss during training. The detailed process is shown in Figure~\ref{fig:fine-tune}. 
\subsubsection{Training Details.} 
We test and tune {\model} on WN18RR dataset. 
During tuning, we construct $10$ negative triples by randomly replacing $h$ or $t$ in $(h,r,t)$. Training batch size is set to $16$ and optimizer is Adam \cite{adam} with learning rate $0.0001$.

\begin{table}[]
    \centering
    \small 
     \caption{Results of triple classification on WN18RR\cite{rotate}. Columns with gray background  are more important.}
     \vspace{-3mm}
    \begin{tabular}{l|>{\columncolor{light-gray}}c| c c >{\columncolor{light-gray}}c}
    \toprule
          & \textbf{Acc.} & \textbf{Precision} & \textbf{Recall} & \textbf{F1}  \\
         \midrule
        TransE \cite{transe} &  88.35 & 93.45 & 82.48 & 87.62  \\ 
        RotatE \cite{rotate} & 88.26 & 93.03 & 82.71 & 87.57 \\
        ComplEx \cite{complex} & 85.07 & \textbf{96.73} & 72.59 & 82.94 \\
        \midrule
        \textit{\textbf{\model}}
         & \textbf{89.21} & {85.56} & \textbf{94.32} & \textbf{89.73} \\
        \textit{KGT-finetune} 
        &  {87.48} & {83.02} & {94.22} & {88.27} \\
        \textit{KGT-scratch} & 67.02 & 67.91 & 64.55 & 66.19 \\
        \bottomrule
    \end{tabular}
    \label{tab:triple-classification}
    \vspace{-3mm}
\end{table}

\subsubsection{Results Analysis.}
We compare results of {\model} to three commonly used KG embedding methods, TransE \cite{transe}, ComplEx \cite{complex}, and RotatE \cite{rotate}, which has proved to be powerful at KG predictions. We evaluate models on 4 classification metrics, including Accuracy(Acc.), Precision, Recall, and F1 score, among which accuracy and F1 are more important metrics. 
Table~\ref{tab:triple-classification} shows results. 

Compared to powerful KGEs, pretrained \textit{{\model}} outperforms them and overall achieves the best triple classification results, especially on metrics of Accuracy and F1 score. {\model} is better at recalling positive triples than KGEs and has a reasonable precision on predicted positive triples, thus resulting in the best F1 score. These results prove that the pretrained {\model} is applicable and effective for in-KG task triple classification. 

Compared to \textit{\model} with $\theta_{\mathcal{M}}$ frozen, overall results of \textit{{KGT}-finetune} is worse, whose averaged accuracy and F1 is $88.27$ while \textit{\model} gives $89.73$, showing the frozen pretrained {\model} layer introduce better local minima during tuning which is beyond what could be reached based on task data only. This is further supported by results of \textit{{KGT}-scratch}, which is significantly worse than \textit{\model}. 

\subsection{Task2: Zero-shot Image Classification}
Zero-shot(ZS) image classification is a challenging task to train models on samples with \textit{seen classes} while test models on samples with \textit{unseen classes} that have no training samples. KGs could be used as auxiliary information for augmenting ZSL tasks\cite{ontozsl,gcnz}.  
\subsubsection{{\model} Implementation.} 
Applying pretrained {\model} to this task, 
we frame the task as outputting the matching score between the input image $I$ and target class $C$. $\mathcal{G}_{task}$ is the class $C$ centered sub-graph. $\mathcal{P}_{task} = [[T]\; [C]\; [M]\; [V]]$ where $[C]$ is the class token sharing the embedding with entity $C \in \mathcal{G}_{task}$; $[V]$ is the token for image $I$ whose representation comes from a vision encoder, e.g. ResNet\cite{resnet}.  
We take out $s_{[B]}^m$ and $s_{[V]}^m$ and project them through an MLP layer, and then calculate the cosine similarity between them as the matching score of the input image $I$ and class $C$. Considering that KG is used to augment unseen classes in this task while not inverse, after generating neighborship matrix $M$ following Equation (\ref{matrix}), 
we specifically set values in the columns corresponding to the prompt token $[T], [C], [M]$ and $[V]$ in $M$ as $0$. That is, $\mathcal{G}_{task}$ can directly affect the learning of the prompt representation, but not vice versa. We applied Binary Cross Entropy loss to encourage the cosine similarity to be large for positive pairs and small for negative pairs. The overall process is shown in Figure~\ref{fig:fine-tune}.

\subsubsection{Training Details.}
Experiments are conducted on AwA-KG \cite{k-zsl}, a benchmark includes samples from AwA dataset \cite{awa} and a basic KG $\mathcal{G}_{task}$ containing hierarchies and attribute annotations of $40$ seen classes and $10$ unseen classes. During tuning, we first train the new embedding table on $\mathcal{G}_{task}$ with three pretraining tasks and then tune the model on the image classification task. 
We use pretrained ResNet \cite{resnet} as vision encoder encoding image $I$ into a $2048$ dimensional vector and transform the image vector into $768$ dimension through a trainable transformation matrix as the representation for token $[V]$. 
We tune the model with batch size set to $24$ and Adam \cite{adam}, whose initial rate is set at $0.0001$.
During tuning, we froze the parameters in ResNet for simplicity.  

\begin{table}[]
    \centering
    \small
    \caption{Results of ZS image classification on AwA-KG\cite{k-zsl}}
    \vspace{-3mm}
    \begin{tabular}{l|>{\columncolor{light-gray}}c| c c >{\columncolor{light-gray}}c}
    \toprule
          & \textbf{T1} & \textbf{S} & \textbf{U} & \textbf{H}    \\
         \midrule
        DeViSE\cite{devise} & 43.24 & \textbf{86.44} & 6.40 & 11.91  \\ 
        GCNZ\cite{gcnz}  & 62.98 & 75.59 & 20.28 & 31.98 \\
        OntoZSL\cite{ontozsl} & 62.65 & 59.59 & 50.58 & 54.71 \\
        \midrule
        \textit{\textbf{\model}}  & \textbf{66.26} & 
 {61.13} & \textbf{55.14} & \textbf{57.98} \\
        \textit{KGT-finetune} 
         & {63.59} & {63.61} & {50.08} & {56.04} \\
        \textit{KGT-scratch} & 58.96 & 56.48 & 47.93 & 51.85 \\
        \bottomrule
    \end{tabular}
    \label{tab:zsl}
    \vspace{-3mm}
\end{table}
\subsubsection{Results Analysis.}
In Table \ref{tab:zsl}, we compare the {\model} with 3 zero-shot learning (ZSL) methods supporting auxiliary KGs. We evaluate our methods and baselines on both conventional ZSL and general ZSL tasks. In conventional ZSL tasks, only images with unseen classes are tested, and only unseen classes are targeted to be classified into, which means we know the test image belongs to unseen classes. In general ZSL tasks, images with seen and unseen classes are tested. Since we do not know the true class of the image is seen or unseen, the candidate classes to be classified into include all seen and unseen classes. Thus general ZSL is a more challenging task. For conventional ZSL, the class-balanced accuracy is reported (T1). For general ZSL, the class-balanced accuracy on seen classes (S), unseen classes (U), and hybrid metrics (H) are reported, where $H = \frac{2(S*U)}{S+U}$. Overall, the H value is a more important metric for general ZSL. The results are shown in Table~\ref{tab:zsl}. 

Compared to baselines, the pretrained \textit{{\model}} achieves the best results on metrics T1, U, and H and overall performs the best.  
On S metrics, DeViSE presents the best results showing that it is significantly biased to seen classes with limited predictive power on unseen classes. 
Comparing different settings of pretrained \textit{{\model}} with $\theta_{\mathcal{M}}$ frozen, \textit{{KGT}-finetune} achieves slightly worse results than \textit{\model} and \textit{{KGT}-scratch} present significantly worse results, which are consistent to the results on triple classification (Table \ref{tab:triple-classification}).
In summary, Table \ref{tab:zsl} shows the pretrained 
\textit{{\model}} is applicable and effective in zero-shot image classification. It successfully builds semantic connections between seen and unseen classes, generalizes to unseen classes during test, and has a better capability of generalizing predictive power on seen classes to unseen classes than baselines.

\begin{table*}[]
    \centering
    \caption{Ablation Study.}
    \vspace{-3mm}
    \small
    \begin{tabular}{c| c c c c | c c c c | c c }
    \toprule
     & \multicolumn{4}{c|}{Triple Classification} 
     & \multicolumn{4}{c|}{Zero-shot Image Classification}
     & \multicolumn{2}{c}{Question Answering}
     \\
     \cline{2-11} 
     & Acc. & Precision & Recall & F1 & T1 & S & U & H & IHdev & IHtest \\
     \hline 
      KGTransformer &  {89.20} &  {85.56} &  {94.32} &  {89.73} &  {66.26} &  {61.13} &  {55.14} &  {57.98} &  {77.64} &  {74.13} \\
      - MEM  &  {86.85} &  {82.15} &  {94.16} &  {87.74} &  {63.44} &  {61.49} &  {52.77} &  {56.80} &  {75.84} &  {72.60}\\
      - MRM  &  {84.91} &  {79.42} &  {94.22} &  {86.19} &  {62.36} &  {66.41} &  {47.97} &  {55.70}  & {74.45} &  {71.47} \\
      - EPM  &  {87.78} &  {83.54} &  {94.10} &  {88.51} &  {65.73} &  {64.13} &  {52.50} &  {57.74} & {76.33} & {72.84}  \\
      -$M$ & 74.15 & 67.43 & 93.42 & 78.33 & 61.47 & 59.98 & 36.33 & 45.25 & 74.12 & 69.70 \\
    \bottomrule
    \end{tabular}

    \label{tab:ablation}
\end{table*}
\begin{figure*}
    \centering
\includegraphics[width=0.8\linewidth]{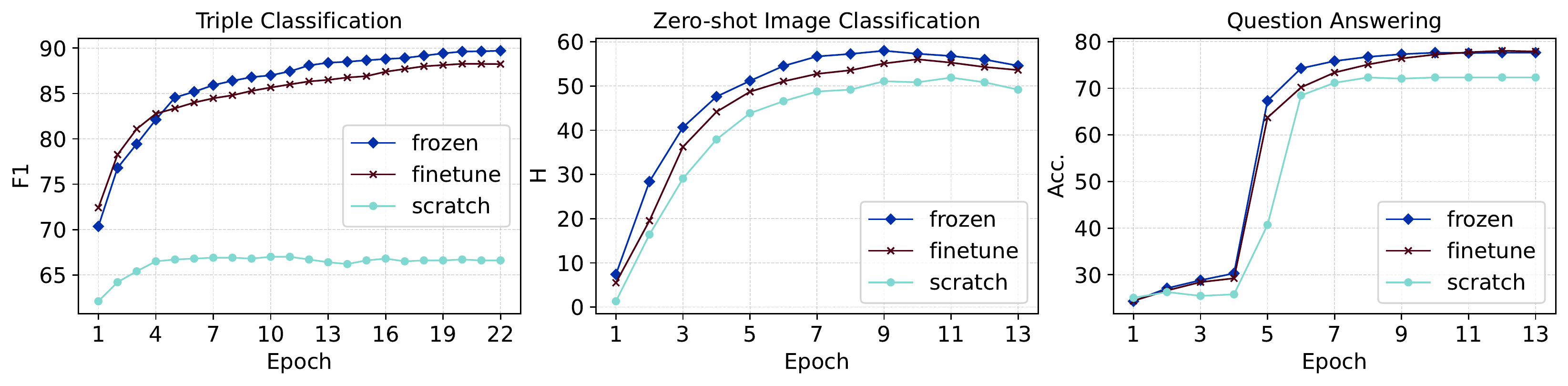}
\vspace{-3mm}
    \caption{Prediction results of three settings of {\model} on three tasks during training.}
    \label{fig:train}
    \vspace{-3mm}
\end{figure*}
\begin{table}[t]
\small 
\caption{Accuracy of QA on CommonsenseQA. 
}
\vspace{-3mm}
\begin{tabular}{lcc}
\toprule
& \textbf{IHdev} & \textbf{IHtest} \\
\midrule
RoBERTa-Large \cite{roberta} & 73.07 & 68.69 \\
RoBERTa-Large \cite{roberta}(ours) &  72.24 & 68.49 \\
\midrule
+ GconAttn \cite{DBLP:conf/aaai/WangKMYTACFMMW19} & 72.61 & 68.59 \\
+ KagNet \cite{DBLP:conf/emnlp/LinCCR19}   & 73.47 & 69.01 \\
+ MHGRN \cite{DBLP:conf/emnlp/FengCLWYR20}   & 74.45 & 71.11 \\
+ QA-GNN \cite{qagnn}   & 76.54 & 73.41 \\
\midrule
+ \textit{\textbf{\model}} 
  & {77.64} & {74.13} \\ 
+ \textit{KGT-finetune} 
 & \textbf{78.05} & \textbf{74.21}\\ 
+ \textit{KGT-scratch}  & 72.32 & 69.22\\ 
\bottomrule
\end{tabular}
\label{tab:qa}
\vspace{-4mm}
\end{table}

\subsection{Task 3: Question Answering (QA)} 
QA task is to select the correct answer given a natural language question. It is a challenging task requiring complex reasoning over constraints stated in the question consistent with the relevant knowledge in the world. 
KGs are used to provide commonsense background knowledge for this challenging task\cite{qagnn}.
\subsubsection{{\model} Implementation.}
Applying {\model} to QA, we make {\model} to predict the likelihood of the input question-choice pair being correct. Given a question-choice pair $qc$, we extract  $\mathcal{G}_{task}$ from ConceptNet\cite{conceptnet} according to a set of keywords $\mathcal{W}$ in $qc$ following \cite{qagnn}. $\mathcal{P}_{task} = [[T]\; [M] \; [M]\; [Q]]$ where $[Q]$ is the token for $qc$ whose representation $R_{qc}$ come from a language encoder, such as BERT\cite{bert}, RoBERTa\cite{roberta}. 
Following \cite{qagnn}, in the task layer, we first concatenate $s^m_{[B]}$, $s^m_{[Q]}$ and $R_{qc}$ and then input them into an MLP layer to predict the likelihood of the $qc$. The overall process is shown in Figure~\ref{fig:fine-tune}. 
\subsubsection{Training Details.}
We experiment on CommonsenQA benchmark \cite{commonsenseqa}, a 5-way multiple-choice QA task.
We report the main results on the in-house (IH) data splits and official test set as did in \cite{qagnn}. We apply  RoBERTa-large \cite{roberta} as language encoder with the first $1024$ dimensional hidden state in the sequence as $R_{qc}$. We transform $R_{qc}$ into a 768-dimensional one through a transformation matrix. 
We regard $\mathcal{P}_{task}$ and a triple in $\mathcal{G}_{task}$ containing words in $\mathcal{W}$ as related during the construction of $M$. We tune the model with batch size set to $128$ 
and Adam\cite{adam} whose initial rate set as $0.00001$. In the first 4 tuning epochs, the RoBERTa-large is frozen.

\subsubsection{Results Analysis}
In Table \ref{tab:qa}, we compare {\model} to recently proposed QA methods and report their accuracy on in-house valid and test sets.
All baselines are using RoBERTa-Large\cite{roberta} as language encoder. 

Compared to baselines, \textit{{\model}} achieves the best results. 
Compared to the RoBERTa-large(ours) that was produced in the same experiment environment, and code framework, our model encoding KG related to question-choice pair by the pretrained KGTransformer successfully improves the QA accuracy on both valid and test datasets. 
The pretrained KGTransformer with three settings all achieves better results than Roberta-Large, among which results of fine-tuning are the best. Results of freezing are comparable to fine-tuning and are better than training from scratch. The reason for \textit{KGT-finetune} performs better than \textit{\model} might be that the QA task mainly relies on the quality of the language encoder(RoBERTa-Large), for example, RoBERTa-Large achieves good results on this task, thus fine-tuning $\theta_{\mathcal{M}}$ will make model adapt better to RoBERTa-Large and achieves better results.

\subsection{Model Analysis}
\subsubsection{Ablation Study} 
In Table~\ref{tab:ablation}, we show the results of {\model}  trained without mask entity modeling(-MEM), without mask relation modeling (-MRM), without entity pair modeling (-EPM), and without neighborship matrix (-M), to illustrate how effective they are for pretraining. 
We adopt frozen tuning in this ablation study on three tasks. 
The results show that removing any one of them will make the task results worse, showing they are beneficial to the pretraining of {\model}. 
Among three tasks, MRM contributes the most, and MEM contributes more than EPM. This is reasonable because MRM is the hardest task. Some MEM tasks could be trickily solved by copying entities from the sequence if the masked one is also contained in other triples. Similarly, some EPM tasks could be trickily solved by comparing whether the two entities are the head(tail) entities of the same relation in the subgraph. 
Without neighborship matrix, the results of three tasks are significantly worse than KGTransformer and KGT-finetune, and slightly better or worse than KGT-scratch. And we find that without the matrix, the pretraining loss did not significantly change and the pretraining model didn't converge. This proves that explicitly telling model the graph structure of KGs is effective for pretraining. 

\subsubsection{Training Details}
In Figure \ref{fig:train}, we show the prediction results of the pretrained {\model} with three settings on three tasks. We observe the same phenomena across three tasks that (1) tuning {\model} with $\theta_{\mathcal{M}}$ frozen leads to good results, which is usually better than or comparable to fine-tuning $\theta_{\mathcal{M}}$, and it is significantly better than training from scratch; (2) tuning {\model} with $\theta_{\mathcal{M}}$ frozen achieves
reasonable results faster. For example, in triple classification, results of $5$ epochs from \textit{\model} are comparable to $10$ epochs from \textit{{KGT}-finetune}. 

Thus we conclude that (1) pretraining {\model} on KGs with diverse structures enables it to learn global graph structure knowledge in KG that could not be sufficiently learned based on only task KG; (2) we recommend tuning the pretrained {\model} with $\theta_{\mathcal{M}}$ frozen to keep and transfer graph structure knowledge learned from $\mathcal{G}_{pre}$ to downstream tasks better and faster.
 
\section{Conclusion and Discussion}
In this paper, we propose a novel KG pretraining model {\model } and prove it is possible to pretrain a model with a general knowledge representation and fusion module in multiple tasks supported by different KGs.
We pretrain KGTransformer on a hybrid KG with diverse graph structures and prompt tuning it uniformly on three typical KG-related tasks. {\model} performs better than specifically designed models across different tasks. 
More importantly, simply applying pretrained KGTransformer off the shelf gives promising results, showing the capability of deep graph structure transfer that the pretrained KGTransformer has. 

Though effective, KGTransformer is a heavier KG model than conventional KGEs. It requires more memory and computation resources. 
In the future, we would like to explore how to apply the pretrained KGTransformer to applications with limited resources, such as mobile applications and edge computing.

\begin{acks}
This work is funded by NSFC91846204/U19B2027. Wen Zhang is supported by Zhejiang Provincial Natural Science Foundation of China (No. LQ23F020017) and Yongjiang Talent Introduction Programme.
\end{acks}

\bibliographystyle{ACM-Reference-Format}
\bibliography{reference.bib}


\begin{thebibliography}{71}


\ifx \showCODEN    \undefined \def \showCODEN     #1{\unskip}     \fi
\ifx \showDOI      \undefined \def \showDOI       #1{#1}\fi
\ifx \showISBNx    \undefined \def \showISBNx     #1{\unskip}     \fi
\ifx \showISBNxiii \undefined \def \showISBNxiii  #1{\unskip}     \fi
\ifx \showISSN     \undefined \def \showISSN      #1{\unskip}     \fi
\ifx \showLCCN     \undefined \def \showLCCN      #1{\unskip}     \fi
\ifx \shownote     \undefined \def \shownote      #1{#1}          \fi
\ifx \showarticletitle \undefined \def \showarticletitle #1{#1}   \fi
\ifx \showURL      \undefined \def \showURL       {\relax}        \fi
\providecommand\bibfield[2]{#2}
\providecommand\bibinfo[2]{#2}
\providecommand\natexlab[1]{#1}
\providecommand\showeprint[2][]{arXiv:#2}

\bibitem[Arnab et~al\mbox{.}(2021)]%
        {vivit}
\bibfield{author}{\bibinfo{person}{Anurag Arnab}, \bibinfo{person}{Mostafa
  Dehghani}, \bibinfo{person}{Georg Heigold}, \bibinfo{person}{Chen Sun},
  \bibinfo{person}{Mario Lucic}, {and} \bibinfo{person}{Cordelia Schmid}.}
  \bibinfo{year}{2021}\natexlab{}.
\newblock \showarticletitle{ViViT: {A} Video Vision Transformer}. In
  \bibinfo{booktitle}{\emph{{ICCV}}}. \bibinfo{publisher}{{IEEE}},
  \bibinfo{pages}{6816--6826}.
\newblock


\bibitem[Bao et~al\mbox{.}(2022)]%
        {beit}
\bibfield{author}{\bibinfo{person}{Hangbo Bao}, \bibinfo{person}{Li Dong},
  \bibinfo{person}{Songhao Piao}, {and} \bibinfo{person}{Furu Wei}.}
  \bibinfo{year}{2022}\natexlab{}.
\newblock \showarticletitle{BEiT: {BERT} Pre-Training of Image Transformers}.
  In \bibinfo{booktitle}{\emph{{ICLR}}}. \bibinfo{publisher}{OpenReview.net}.
\newblock


\bibitem[Bordes et~al\mbox{.}(2013)]%
        {transe}
\bibfield{author}{\bibinfo{person}{Antoine Bordes}, \bibinfo{person}{Nicolas
  Usunier}, \bibinfo{person}{Alberto Garc{\'{\i}}a{-}Dur{\'{a}}n},
  \bibinfo{person}{Jason Weston}, {and} \bibinfo{person}{Oksana Yakhnenko}.}
  \bibinfo{year}{2013}\natexlab{}.
\newblock \showarticletitle{Translating Embeddings for Modeling
  Multi-relational Data}. In \bibinfo{booktitle}{\emph{{NIPS}}}.
  \bibinfo{pages}{2787--2795}.
\newblock


\bibitem[Carlson et~al\mbox{.}(2010)]%
        {nell}
\bibfield{author}{\bibinfo{person}{Andrew Carlson}, \bibinfo{person}{Justin
  Betteridge}, \bibinfo{person}{Bryan Kisiel}, \bibinfo{person}{Burr Settles},
  \bibinfo{person}{Estevam R.~Hruschka Jr.}, {and} \bibinfo{person}{Tom~M.
  Mitchell}.} \bibinfo{year}{2010}\natexlab{}.
\newblock \showarticletitle{Toward an Architecture for Never-Ending Language
  Learning}. In \bibinfo{booktitle}{\emph{{AAAI}}}. \bibinfo{publisher}{{AAAI}
  Press}.
\newblock


\bibitem[Chen et~al\mbox{.}(2021)]%
        {DBLP:conf/emnlp/ChenLG0ZJ21}
\bibfield{author}{\bibinfo{person}{Sanxing Chen}, \bibinfo{person}{Xiaodong
  Liu}, \bibinfo{person}{Jianfeng Gao}, \bibinfo{person}{Jian Jiao},
  \bibinfo{person}{Ruofei Zhang}, {and} \bibinfo{person}{Yangfeng Ji}.}
  \bibinfo{year}{2021}\natexlab{}.
\newblock \showarticletitle{HittER: Hierarchical Transformers for Knowledge
  Graph Embeddings}. In \bibinfo{booktitle}{\emph{{EMNLP} {(1)}}}.
  \bibinfo{publisher}{Association for Computational Linguistics},
  \bibinfo{pages}{10395--10407}.
\newblock


\bibitem[Chen et~al\mbox{.}(2022)]%
        {DBLP:conf/sigir/ChenZLDTXHSC22}
\bibfield{author}{\bibinfo{person}{Xiang Chen}, \bibinfo{person}{Ningyu Zhang},
  \bibinfo{person}{Lei Li}, \bibinfo{person}{Shumin Deng},
  \bibinfo{person}{Chuanqi Tan}, \bibinfo{person}{Changliang Xu},
  \bibinfo{person}{Fei Huang}, \bibinfo{person}{Luo Si}, {and}
  \bibinfo{person}{Huajun Chen}.} \bibinfo{year}{2022}\natexlab{}.
\newblock \showarticletitle{Hybrid Transformer with Multi-level Fusion for
  Multimodal Knowledge Graph Completion}. In
  \bibinfo{booktitle}{\emph{{SIGIR}}}. \bibinfo{publisher}{{ACM}},
  \bibinfo{pages}{904--915}.
\newblock


\bibitem[Devlin et~al\mbox{.}(2019)]%
        {bert}
\bibfield{author}{\bibinfo{person}{Jacob Devlin}, \bibinfo{person}{Ming{-}Wei
  Chang}, \bibinfo{person}{Kenton Lee}, {and} \bibinfo{person}{Kristina
  Toutanova}.} \bibinfo{year}{2019}\natexlab{}.
\newblock \showarticletitle{{BERT:} Pre-training of Deep Bidirectional
  Transformers for Language Understanding}. In
  \bibinfo{booktitle}{\emph{{NAACL-HLT} {(1)}}}.
  \bibinfo{publisher}{Association for Computational Linguistics},
  \bibinfo{pages}{4171--4186}.
\newblock


\bibitem[Dosovitskiy et~al\mbox{.}(2021)]%
        {vit}
\bibfield{author}{\bibinfo{person}{Alexey Dosovitskiy}, \bibinfo{person}{Lucas
  Beyer}, \bibinfo{person}{Alexander Kolesnikov}, \bibinfo{person}{Dirk
  Weissenborn}, \bibinfo{person}{Xiaohua Zhai}, \bibinfo{person}{Thomas
  Unterthiner}, \bibinfo{person}{Mostafa Dehghani}, \bibinfo{person}{Matthias
  Minderer}, \bibinfo{person}{Georg Heigold}, \bibinfo{person}{Sylvain Gelly},
  \bibinfo{person}{Jakob Uszkoreit}, {and} \bibinfo{person}{Neil Houlsby}.}
  \bibinfo{year}{2021}\natexlab{}.
\newblock \showarticletitle{An Image is Worth 16x16 Words: Transformers for
  Image Recognition at Scale}. In \bibinfo{booktitle}{\emph{{ICLR}}}.
  \bibinfo{publisher}{OpenReview.net}.
\newblock


\bibitem[Feng et~al\mbox{.}(2020)]%
        {DBLP:conf/emnlp/FengCLWYR20}
\bibfield{author}{\bibinfo{person}{Yanlin Feng}, \bibinfo{person}{Xinyue Chen},
  \bibinfo{person}{Bill~Yuchen Lin}, \bibinfo{person}{Peifeng Wang},
  \bibinfo{person}{Jun Yan}, {and} \bibinfo{person}{Xiang Ren}.}
  \bibinfo{year}{2020}\natexlab{}.
\newblock \showarticletitle{Scalable Multi-Hop Relational Reasoning for
  Knowledge-Aware Question Answering}. In \bibinfo{booktitle}{\emph{{EMNLP}
  {(1)}}}. \bibinfo{publisher}{Association for Computational Linguistics},
  \bibinfo{pages}{1295--1309}.
\newblock


\bibitem[Frome et~al\mbox{.}(2013)]%
        {devise}
\bibfield{author}{\bibinfo{person}{Andrea Frome}, \bibinfo{person}{Gregory~S.
  Corrado}, \bibinfo{person}{Jonathon Shlens}, \bibinfo{person}{Samy Bengio},
  \bibinfo{person}{Jeffrey Dean}, \bibinfo{person}{Marc'Aurelio Ranzato}, {and}
  \bibinfo{person}{Tom{\'{a}}s Mikolov}.} \bibinfo{year}{2013}\natexlab{}.
\newblock \showarticletitle{DeViSE: {A} Deep Visual-Semantic Embedding Model}.
  In \bibinfo{booktitle}{\emph{{NIPS}}}. \bibinfo{pages}{2121--2129}.
\newblock


\bibitem[Gao et~al\mbox{.}(2020)]%
        {mm-pretrain2}
\bibfield{author}{\bibinfo{person}{Dehong Gao}, \bibinfo{person}{Linbo Jin},
  \bibinfo{person}{Ben Chen}, \bibinfo{person}{Minghui Qiu},
  \bibinfo{person}{Peng Li}, \bibinfo{person}{Yi Wei}, \bibinfo{person}{Yi Hu},
  {and} \bibinfo{person}{Hao Wang}.} \bibinfo{year}{2020}\natexlab{}.
\newblock \showarticletitle{FashionBERT: Text and Image Matching with Adaptive
  Loss for Cross-modal Retrieval}. In \bibinfo{booktitle}{\emph{{SIGIR}}}.
  \bibinfo{publisher}{{ACM}}, \bibinfo{pages}{2251--2260}.
\newblock


\bibitem[Gao et~al\mbox{.}(2021)]%
        {DBLP:conf/acl/GaoFC20}
\bibfield{author}{\bibinfo{person}{Tianyu Gao}, \bibinfo{person}{Adam Fisch},
  {and} \bibinfo{person}{Danqi Chen}.} \bibinfo{year}{2021}\natexlab{}.
\newblock \showarticletitle{Making Pre-trained Language Models Better Few-shot
  Learners}. In \bibinfo{booktitle}{\emph{{ACL/IJCNLP} {(1)}}}.
  \bibinfo{publisher}{Association for Computational Linguistics},
  \bibinfo{pages}{3816--3830}.
\newblock


\bibitem[Gard{\`{e}}res et~al\mbox{.}(2020)]%
        {DBLP:conf/emnlp/GarderesZAL20}
\bibfield{author}{\bibinfo{person}{Fran{\c{c}}ois Gard{\`{e}}res},
  \bibinfo{person}{Maryam Ziaeefard}, \bibinfo{person}{Baptiste Abeloos}, {and}
  \bibinfo{person}{Freddy L{\'{e}}cu{\'{e}}}.} \bibinfo{year}{2020}\natexlab{}.
\newblock \showarticletitle{ConceptBert: Concept-Aware Representation for
  Visual Question Answering}. In \bibinfo{booktitle}{\emph{{EMNLP} (Findings)}}
  \emph{(\bibinfo{series}{Findings of {ACL}}, Vol.~\bibinfo{volume}{{EMNLP}
  2020})}. \bibinfo{publisher}{Association for Computational Linguistics},
  \bibinfo{pages}{489--498}.
\newblock


\bibitem[Geng et~al\mbox{.}(2021a)]%
        {ontozsl}
\bibfield{author}{\bibinfo{person}{Yuxia Geng}, \bibinfo{person}{Jiaoyan Chen},
  \bibinfo{person}{Zhuo Chen}, \bibinfo{person}{Jeff~Z. Pan},
  \bibinfo{person}{Zhiquan Ye}, \bibinfo{person}{Zonggang Yuan},
  \bibinfo{person}{Yantao Jia}, {and} \bibinfo{person}{Huajun Chen}.}
  \bibinfo{year}{2021}\natexlab{a}.
\newblock \showarticletitle{OntoZSL: Ontology-enhanced Zero-shot Learning}. In
  \bibinfo{booktitle}{\emph{{WWW}}}. \bibinfo{publisher}{{ACM} / {IW3C2}},
  \bibinfo{pages}{3325--3336}.
\newblock


\bibitem[Geng et~al\mbox{.}(2021b)]%
        {k-zsl}
\bibfield{author}{\bibinfo{person}{Yuxia Geng}, \bibinfo{person}{Jiaoyan Chen},
  \bibinfo{person}{Zhuo Chen}, \bibinfo{person}{Jeff~Z Pan},
  \bibinfo{person}{Zonggang Yuan}, {and} \bibinfo{person}{Huajun Chen}.}
  \bibinfo{year}{2021}\natexlab{b}.
\newblock \showarticletitle{K-ZSL: resources for knowledge-driven zero-shot
  learning}.
\newblock \bibinfo{journal}{\emph{arXiv preprint arXiv:2106.15047}}
  (\bibinfo{year}{2021}).
\newblock


\bibitem[Geng et~al\mbox{.}(2022)]%
        {DOZSL}
\bibfield{author}{\bibinfo{person}{Yuxia Geng}, \bibinfo{person}{Jiaoyan Chen},
  \bibinfo{person}{Wen Zhang}, \bibinfo{person}{Yajing Xu},
  \bibinfo{person}{Zhuo Chen}, \bibinfo{person}{Jeff~Z. Pan},
  \bibinfo{person}{Yufeng Huang}, \bibinfo{person}{Feiyu Xiong}, {and}
  \bibinfo{person}{Huajun Chen}.} \bibinfo{year}{2022}\natexlab{}.
\newblock \showarticletitle{Disentangled Ontology Embedding for Zero-shot
  Learning}. In \bibinfo{booktitle}{\emph{{KDD}}}. \bibinfo{publisher}{{ACM}},
  \bibinfo{pages}{443--453}.
\newblock


\bibitem[He et~al\mbox{.}(2016)]%
        {resnet}
\bibfield{author}{\bibinfo{person}{Kaiming He}, \bibinfo{person}{Xiangyu
  Zhang}, \bibinfo{person}{Shaoqing Ren}, {and} \bibinfo{person}{Jian Sun}.}
  \bibinfo{year}{2016}\natexlab{}.
\newblock \showarticletitle{Deep Residual Learning for Image Recognition}. In
  \bibinfo{booktitle}{\emph{{CVPR}}}. \bibinfo{publisher}{{IEEE} Computer
  Society}, \bibinfo{pages}{770--778}.
\newblock


\bibitem[Herzig et~al\mbox{.}(2020)]%
        {table2}
\bibfield{author}{\bibinfo{person}{Jonathan Herzig},
  \bibinfo{person}{Pawel~Krzysztof Nowak}, \bibinfo{person}{Thomas
  M{\"{u}}ller}, \bibinfo{person}{Francesco Piccinno}, {and}
  \bibinfo{person}{Julian~Martin Eisenschlos}.}
  \bibinfo{year}{2020}\natexlab{}.
\newblock \showarticletitle{TaPas: Weakly Supervised Table Parsing via
  Pre-training}. In \bibinfo{booktitle}{\emph{{ACL}}}.
  \bibinfo{publisher}{Association for Computational Linguistics},
  \bibinfo{pages}{4320--4333}.
\newblock


\bibitem[Hu et~al\mbox{.}(2022)]%
        {DBLP:conf/emnlp/HuGXLP22}
\bibfield{author}{\bibinfo{person}{Zhiwei Hu}, \bibinfo{person}{V{\'{\i}}ctor
  Guti{\'{e}}rrez{-}Basulto}, \bibinfo{person}{Zhiliang Xiang},
  \bibinfo{person}{Ru Li}, {and} \bibinfo{person}{Jeff~Z. Pan}.}
  \bibinfo{year}{2022}\natexlab{}.
\newblock \showarticletitle{Transformer-based Entity Typing in Knowledge
  Graphs}. In \bibinfo{booktitle}{\emph{{EMNLP}}}.
  \bibinfo{publisher}{Association for Computational Linguistics},
  \bibinfo{pages}{5988--6001}.
\newblock


\bibitem[Huang et~al\mbox{.}(2019)]%
        {QA-KG}
\bibfield{author}{\bibinfo{person}{Xiao Huang}, \bibinfo{person}{Jingyuan
  Zhang}, \bibinfo{person}{Dingcheng Li}, {and} \bibinfo{person}{Ping Li}.}
  \bibinfo{year}{2019}\natexlab{}.
\newblock \showarticletitle{Knowledge Graph Embedding Based Question
  Answering}. In \bibinfo{booktitle}{\emph{{WSDM}}}.
  \bibinfo{publisher}{{ACM}}, \bibinfo{pages}{105--113}.
\newblock


\bibitem[Kim et~al\mbox{.}(2020)]%
        {DBLP:conf/coling/KimHKS20}
\bibfield{author}{\bibinfo{person}{Bosung Kim}, \bibinfo{person}{Taesuk Hong},
  \bibinfo{person}{Youngjoong Ko}, {and} \bibinfo{person}{Jungyun Seo}.}
  \bibinfo{year}{2020}\natexlab{}.
\newblock \showarticletitle{Multi-Task Learning for Knowledge Graph Completion
  with Pre-trained Language Models}. In \bibinfo{booktitle}{\emph{{COLING}}}.
  \bibinfo{publisher}{International Committee on Computational Linguistics},
  \bibinfo{pages}{1737--1743}.
\newblock


\bibitem[Kingma and Ba(2014)]%
        {adam}
\bibfield{author}{\bibinfo{person}{Diederik~P Kingma} {and}
  \bibinfo{person}{Jimmy Ba}.} \bibinfo{year}{2014}\natexlab{}.
\newblock \showarticletitle{Adam: A method for stochastic optimization}.
\newblock \bibinfo{journal}{\emph{arXiv preprint arXiv:1412.6980}}
  (\bibinfo{year}{2014}).
\newblock


\bibitem[Koncel{-}Kedziorski et~al\mbox{.}(2019)]%
        {DBLP:conf/naacl/Koncel-Kedziorski19}
\bibfield{author}{\bibinfo{person}{Rik Koncel{-}Kedziorski},
  \bibinfo{person}{Dhanush Bekal}, \bibinfo{person}{Yi Luan},
  \bibinfo{person}{Mirella Lapata}, {and} \bibinfo{person}{Hannaneh
  Hajishirzi}.} \bibinfo{year}{2019}\natexlab{}.
\newblock \showarticletitle{Text Generation from Knowledge Graphs with Graph
  Transformers}. In \bibinfo{booktitle}{\emph{{NAACL-HLT} {(1)}}}.
  \bibinfo{publisher}{Association for Computational Linguistics},
  \bibinfo{pages}{2284--2293}.
\newblock


\bibitem[Lester et~al\mbox{.}(2021)]%
        {prompttuning}
\bibfield{author}{\bibinfo{person}{Brian Lester}, \bibinfo{person}{Rami
  Al{-}Rfou}, {and} \bibinfo{person}{Noah Constant}.}
  \bibinfo{year}{2021}\natexlab{}.
\newblock \showarticletitle{The Power of Scale for Parameter-Efficient Prompt
  Tuning}. In \bibinfo{booktitle}{\emph{{EMNLP} {(1)}}}.
  \bibinfo{publisher}{Association for Computational Linguistics},
  \bibinfo{pages}{3045--3059}.
\newblock


\bibitem[Lewis et~al\mbox{.}(2020)]%
        {bart}
\bibfield{author}{\bibinfo{person}{Mike Lewis}, \bibinfo{person}{Yinhan Liu},
  \bibinfo{person}{Naman Goyal}, \bibinfo{person}{Marjan Ghazvininejad},
  \bibinfo{person}{Abdelrahman Mohamed}, \bibinfo{person}{Omer Levy},
  \bibinfo{person}{Veselin Stoyanov}, {and} \bibinfo{person}{Luke
  Zettlemoyer}.} \bibinfo{year}{2020}\natexlab{}.
\newblock \showarticletitle{{BART:} Denoising Sequence-to-Sequence Pre-training
  for Natural Language Generation, Translation, and Comprehension}. In
  \bibinfo{booktitle}{\emph{{ACL}}}. \bibinfo{publisher}{Association for
  Computational Linguistics}, \bibinfo{pages}{7871--7880}.
\newblock


\bibitem[Li et~al\mbox{.}(2022b)]%
        {DBLP:conf/kdd/LiZZ22}
\bibfield{author}{\bibinfo{person}{Han Li}, \bibinfo{person}{Dan Zhao}, {and}
  \bibinfo{person}{Jianyang Zeng}.} \bibinfo{year}{2022}\natexlab{b}.
\newblock \showarticletitle{{KPGT:} Knowledge-Guided Pre-training of Graph
  Transformer for Molecular Property Prediction}. In
  \bibinfo{booktitle}{\emph{{KDD}}}. \bibinfo{publisher}{{ACM}},
  \bibinfo{pages}{857--867}.
\newblock


\bibitem[Li et~al\mbox{.}(2022a)]%
        {hran}
\bibfield{author}{\bibinfo{person}{Zhifei Li}, \bibinfo{person}{Hai Liu},
  \bibinfo{person}{Zhaoli Zhang}, \bibinfo{person}{Tingting Liu}, {and}
  \bibinfo{person}{Neal~N. Xiong}.} \bibinfo{year}{2022}\natexlab{a}.
\newblock \showarticletitle{Learning Knowledge Graph Embedding With
  Heterogeneous Relation Attention Networks}.
\newblock \bibinfo{journal}{\emph{{IEEE} Trans. Neural Networks Learn. Syst.}}
  \bibinfo{volume}{33}, \bibinfo{number}{8} (\bibinfo{year}{2022}),
  \bibinfo{pages}{3961--3973}.
\newblock


\bibitem[Lin et~al\mbox{.}(2019)]%
        {DBLP:conf/emnlp/LinCCR19}
\bibfield{author}{\bibinfo{person}{Bill~Yuchen Lin}, \bibinfo{person}{Xinyue
  Chen}, \bibinfo{person}{Jamin Chen}, {and} \bibinfo{person}{Xiang Ren}.}
  \bibinfo{year}{2019}\natexlab{}.
\newblock \showarticletitle{KagNet: Knowledge-Aware Graph Networks for
  Commonsense Reasoning}. In \bibinfo{booktitle}{\emph{{EMNLP/IJCNLP} {(1)}}}.
  \bibinfo{publisher}{Association for Computational Linguistics},
  \bibinfo{pages}{2829--2839}.
\newblock


\bibitem[Lin et~al\mbox{.}(2022)]%
        {inductive-sigir}
\bibfield{author}{\bibinfo{person}{Qika Lin}, \bibinfo{person}{Jun Liu},
  \bibinfo{person}{Fangzhi Xu}, \bibinfo{person}{Yudai Pan},
  \bibinfo{person}{Yifan Zhu}, \bibinfo{person}{Lingling Zhang}, {and}
  \bibinfo{person}{Tianzhe Zhao}.} \bibinfo{year}{2022}\natexlab{}.
\newblock \showarticletitle{Incorporating Context Graph with Logical Reasoning
  for Inductive Relation Prediction}. In \bibinfo{booktitle}{\emph{{SIGIR}}}.
  \bibinfo{publisher}{{ACM}}, \bibinfo{pages}{893--903}.
\newblock


\bibitem[Lin et~al\mbox{.}(2015)]%
        {transr}
\bibfield{author}{\bibinfo{person}{Yankai Lin}, \bibinfo{person}{Zhiyuan Liu},
  \bibinfo{person}{Maosong Sun}, \bibinfo{person}{Yang Liu}, {and}
  \bibinfo{person}{Xuan Zhu}.} \bibinfo{year}{2015}\natexlab{}.
\newblock \showarticletitle{Learning Entity and Relation Embeddings for
  Knowledge Graph Completion}. In \bibinfo{booktitle}{\emph{{AAAI}}}.
  \bibinfo{publisher}{{AAAI} Press}, \bibinfo{pages}{2181--2187}.
\newblock


\bibitem[Liu et~al\mbox{.}(2022a)]%
        {BiNet}
\bibfield{author}{\bibinfo{person}{Lihui Liu}, \bibinfo{person}{Boxin Du},
  \bibinfo{person}{Jiejun Xu}, \bibinfo{person}{Yinglong Xia}, {and}
  \bibinfo{person}{Hanghang Tong}.} \bibinfo{year}{2022}\natexlab{a}.
\newblock \showarticletitle{Joint Knowledge Graph Completion and Question
  Answering}. In \bibinfo{booktitle}{\emph{{KDD}}}. \bibinfo{publisher}{{ACM}},
  \bibinfo{pages}{1098--1108}.
\newblock


\bibitem[Liu et~al\mbox{.}(2021)]%
        {liu2021pre}
\bibfield{author}{\bibinfo{person}{Pengfei Liu}, \bibinfo{person}{Weizhe Yuan},
  \bibinfo{person}{Jinlan Fu}, \bibinfo{person}{Zhengbao Jiang},
  \bibinfo{person}{Hiroaki Hayashi}, {and} \bibinfo{person}{Graham Neubig}.}
  \bibinfo{year}{2021}\natexlab{}.
\newblock \showarticletitle{Pre-train, prompt, and predict: A systematic survey
  of prompting methods in natural language processing}.
\newblock \bibinfo{journal}{\emph{arXiv preprint arXiv:2107.13586}}
  (\bibinfo{year}{2021}).
\newblock


\bibitem[Liu et~al\mbox{.}(2022b)]%
        {DBLP:conf/kdd/LiuZSCQZ0DT22}
\bibfield{author}{\bibinfo{person}{Xiao Liu}, \bibinfo{person}{Shiyu Zhao},
  \bibinfo{person}{Kai Su}, \bibinfo{person}{Yukuo Cen},
  \bibinfo{person}{Jiezhong Qiu}, \bibinfo{person}{Mengdi Zhang},
  \bibinfo{person}{Wei Wu}, \bibinfo{person}{Yuxiao Dong}, {and}
  \bibinfo{person}{Jie Tang}.} \bibinfo{year}{2022}\natexlab{b}.
\newblock \showarticletitle{Mask and Reason: Pre-Training Knowledge Graph
  Transformers for Complex Logical Queries}. In
  \bibinfo{booktitle}{\emph{{KDD}}}. \bibinfo{publisher}{{ACM}},
  \bibinfo{pages}{1120--1130}.
\newblock


\bibitem[Liu et~al\mbox{.}(2019)]%
        {roberta}
\bibfield{author}{\bibinfo{person}{Yinhan Liu}, \bibinfo{person}{Myle Ott},
  \bibinfo{person}{Naman Goyal}, \bibinfo{person}{Jingfei Du},
  \bibinfo{person}{Mandar Joshi}, \bibinfo{person}{Danqi Chen},
  \bibinfo{person}{Omer Levy}, \bibinfo{person}{Mike Lewis},
  \bibinfo{person}{Luke Zettlemoyer}, {and} \bibinfo{person}{Veselin
  Stoyanov}.} \bibinfo{year}{2019}\natexlab{}.
\newblock \showarticletitle{RoBERTa: {A} Robustly Optimized {BERT} Pretraining
  Approach}.
\newblock \bibinfo{journal}{\emph{CoRR}}  \bibinfo{volume}{abs/1907.11692}
  (\bibinfo{year}{2019}).
\newblock


\bibitem[Lu et~al\mbox{.}(2019)]%
        {vilbert}
\bibfield{author}{\bibinfo{person}{Jiasen Lu}, \bibinfo{person}{Dhruv Batra},
  \bibinfo{person}{Devi Parikh}, {and} \bibinfo{person}{Stefan Lee}.}
  \bibinfo{year}{2019}\natexlab{}.
\newblock \showarticletitle{ViLBERT: Pretraining Task-Agnostic Visiolinguistic
  Representations for Vision-and-Language Tasks}. In
  \bibinfo{booktitle}{\emph{NeurIPS}}. \bibinfo{pages}{13--23}.
\newblock


\bibitem[Mai et~al\mbox{.}(2021)]%
        {compile}
\bibfield{author}{\bibinfo{person}{Sijie Mai}, \bibinfo{person}{Shuangjia
  Zheng}, \bibinfo{person}{Yuedong Yang}, {and} \bibinfo{person}{Haifeng Hu}.}
  \bibinfo{year}{2021}\natexlab{}.
\newblock \showarticletitle{Communicative Message Passing for Inductive
  Relation Reasoning}. In \bibinfo{booktitle}{\emph{{AAAI}}}.
  \bibinfo{publisher}{{AAAI} Press}, \bibinfo{pages}{4294--4302}.
\newblock


\bibitem[Radford et~al\mbox{.}(2018)]%
        {radford2018improving}
\bibfield{author}{\bibinfo{person}{Alec Radford}, \bibinfo{person}{Karthik
  Narasimhan}, \bibinfo{person}{Tim Salimans}, {and} \bibinfo{person}{Ilya
  Sutskever}.} \bibinfo{year}{2018}\natexlab{}.
\newblock \showarticletitle{Improving language understanding by generative
  pre-training}.
\newblock  (\bibinfo{year}{2018}).
\newblock


\bibitem[Raffel et~al\mbox{.}(2020)]%
        {DBLP:journals/jmlr/RaffelSRLNMZLL20}
\bibfield{author}{\bibinfo{person}{Colin Raffel}, \bibinfo{person}{Noam
  Shazeer}, \bibinfo{person}{Adam Roberts}, \bibinfo{person}{Katherine Lee},
  \bibinfo{person}{Sharan Narang}, \bibinfo{person}{Michael Matena},
  \bibinfo{person}{Yanqi Zhou}, \bibinfo{person}{Wei Li}, {and}
  \bibinfo{person}{Peter~J. Liu}.} \bibinfo{year}{2020}\natexlab{}.
\newblock \showarticletitle{Exploring the Limits of Transfer Learning with a
  Unified Text-to-Text Transformer}.
\newblock \bibinfo{journal}{\emph{J. Mach. Learn. Res.}}  \bibinfo{volume}{21}
  (\bibinfo{year}{2020}), \bibinfo{pages}{140:1--140:67}.
\newblock


\bibitem[Rebele et~al\mbox{.}(2016)]%
        {yago}
\bibfield{author}{\bibinfo{person}{Thomas Rebele}, \bibinfo{person}{Fabian~M.
  Suchanek}, \bibinfo{person}{Johannes Hoffart}, \bibinfo{person}{Joanna
  Biega}, \bibinfo{person}{Erdal Kuzey}, {and} \bibinfo{person}{Gerhard
  Weikum}.} \bibinfo{year}{2016}\natexlab{}.
\newblock \showarticletitle{{YAGO:} {A} Multilingual Knowledge Base from
  Wikipedia, Wordnet, and Geonames}. In \bibinfo{booktitle}{\emph{{ISWC}
  {(2)}}} \emph{(\bibinfo{series}{Lecture Notes in Computer Science},
  Vol.~\bibinfo{volume}{9982})}. \bibinfo{pages}{177--185}.
\newblock


\bibitem[Saxena et~al\mbox{.}(2022)]%
        {KGT5}
\bibfield{author}{\bibinfo{person}{Apoorv Saxena}, \bibinfo{person}{Adrian
  Kochsiek}, {and} \bibinfo{person}{Rainer Gemulla}.}
  \bibinfo{year}{2022}\natexlab{}.
\newblock \showarticletitle{Sequence-to-Sequence Knowledge Graph Completion and
  Question Answering}. In \bibinfo{booktitle}{\emph{{ACL} {(1)}}}.
  \bibinfo{publisher}{Association for Computational Linguistics},
  \bibinfo{pages}{2814--2828}.
\newblock


\bibitem[Saxena et~al\mbox{.}(2020)]%
        {embedKGQA}
\bibfield{author}{\bibinfo{person}{Apoorv Saxena}, \bibinfo{person}{Aditay
  Tripathi}, {and} \bibinfo{person}{Partha~P. Talukdar}.}
  \bibinfo{year}{2020}\natexlab{}.
\newblock \showarticletitle{Improving Multi-hop Question Answering over
  Knowledge Graphs using Knowledge Base Embeddings}. In
  \bibinfo{booktitle}{\emph{{ACL}}}. \bibinfo{publisher}{Association for
  Computational Linguistics}, \bibinfo{pages}{4498--4507}.
\newblock


\bibitem[Schlichtkrull et~al\mbox{.}(2018)]%
        {rgcn}
\bibfield{author}{\bibinfo{person}{Michael~Sejr Schlichtkrull},
  \bibinfo{person}{Thomas~N. Kipf}, \bibinfo{person}{Peter Bloem},
  \bibinfo{person}{Rianne van~den Berg}, \bibinfo{person}{Ivan Titov}, {and}
  \bibinfo{person}{Max Welling}.} \bibinfo{year}{2018}\natexlab{}.
\newblock \showarticletitle{Modeling Relational Data with Graph Convolutional
  Networks}. In \bibinfo{booktitle}{\emph{{ESWC}}}
  \emph{(\bibinfo{series}{Lecture Notes in Computer Science},
  Vol.~\bibinfo{volume}{10843})}. \bibinfo{publisher}{Springer},
  \bibinfo{pages}{593--607}.
\newblock


\bibitem[Speer et~al\mbox{.}(2017)]%
        {conceptnet}
\bibfield{author}{\bibinfo{person}{Robyn Speer}, \bibinfo{person}{Joshua Chin},
  {and} \bibinfo{person}{Catherine Havasi}.} \bibinfo{year}{2017}\natexlab{}.
\newblock \showarticletitle{ConceptNet 5.5: An Open Multilingual Graph of
  General Knowledge}. In \bibinfo{booktitle}{\emph{{AAAI}}}.
  \bibinfo{publisher}{{AAAI} Press}, \bibinfo{pages}{4444--4451}.
\newblock


\bibitem[Su et~al\mbox{.}(2020)]%
        {mm-pretrain1}
\bibfield{author}{\bibinfo{person}{Weijie Su}, \bibinfo{person}{Xizhou Zhu},
  \bibinfo{person}{Yue Cao}, \bibinfo{person}{Bin Li}, \bibinfo{person}{Lewei
  Lu}, \bibinfo{person}{Furu Wei}, {and} \bibinfo{person}{Jifeng Dai}.}
  \bibinfo{year}{2020}\natexlab{}.
\newblock \showarticletitle{{VL-BERT:} Pre-training of Generic
  Visual-Linguistic Representations}. In \bibinfo{booktitle}{\emph{{ICLR}}}.
  \bibinfo{publisher}{OpenReview.net}.
\newblock


\bibitem[Sun et~al\mbox{.}(2022)]%
        {DBLP:conf/emnlp/SunGZ0022}
\bibfield{author}{\bibinfo{person}{Haohai Sun}, \bibinfo{person}{Shangyi Geng},
  \bibinfo{person}{Jialun Zhong}, \bibinfo{person}{Han Hu}, {and}
  \bibinfo{person}{Kun He}.} \bibinfo{year}{2022}\natexlab{}.
\newblock \showarticletitle{Graph Hawkes Transformer for Extrapolated Reasoning
  on Temporal Knowledge Graphs}. In \bibinfo{booktitle}{\emph{{EMNLP}}}.
  \bibinfo{publisher}{Association for Computational Linguistics},
  \bibinfo{pages}{7481--7493}.
\newblock


\bibitem[Sun et~al\mbox{.}(2019)]%
        {rotate}
\bibfield{author}{\bibinfo{person}{Zhiqing Sun}, \bibinfo{person}{Zhi{-}Hong
  Deng}, \bibinfo{person}{Jian{-}Yun Nie}, {and} \bibinfo{person}{Jian Tang}.}
  \bibinfo{year}{2019}\natexlab{}.
\newblock \showarticletitle{RotatE: Knowledge Graph Embedding by Relational
  Rotation in Complex Space}. In \bibinfo{booktitle}{\emph{{ICLR} (Poster)}}.
  \bibinfo{publisher}{OpenReview.net}.
\newblock


\bibitem[Sun et~al\mbox{.}(2018)]%
        {DBLP:conf/ijcai/SunHZQ18}
\bibfield{author}{\bibinfo{person}{Zequn Sun}, \bibinfo{person}{Wei Hu},
  \bibinfo{person}{Qingheng Zhang}, {and} \bibinfo{person}{Yuzhong Qu}.}
  \bibinfo{year}{2018}\natexlab{}.
\newblock \showarticletitle{Bootstrapping Entity Alignment with Knowledge Graph
  Embedding}. In \bibinfo{booktitle}{\emph{{IJCAI}}}.
  \bibinfo{publisher}{ijcai.org}, \bibinfo{pages}{4396--4402}.
\newblock


\bibitem[Talmor et~al\mbox{.}(2019)]%
        {commonsenseqa}
\bibfield{author}{\bibinfo{person}{Alon Talmor}, \bibinfo{person}{Jonathan
  Herzig}, \bibinfo{person}{Nicholas Lourie}, {and} \bibinfo{person}{Jonathan
  Berant}.} \bibinfo{year}{2019}\natexlab{}.
\newblock \showarticletitle{CommonsenseQA: {A} Question Answering Challenge
  Targeting Commonsense Knowledge}. In \bibinfo{booktitle}{\emph{{NAACL-HLT}
  {(1)}}}. \bibinfo{publisher}{Association for Computational Linguistics},
  \bibinfo{pages}{4149--4158}.
\newblock


\bibitem[Teru et~al\mbox{.}(2020)]%
        {grail}
\bibfield{author}{\bibinfo{person}{Komal~K. Teru}, \bibinfo{person}{Etienne~G.
  Denis}, {and} \bibinfo{person}{William~L. Hamilton}.}
  \bibinfo{year}{2020}\natexlab{}.
\newblock \showarticletitle{Inductive Relation Prediction by Subgraph
  Reasoning}. In \bibinfo{booktitle}{\emph{{ICML}}}
  \emph{(\bibinfo{series}{Proceedings of Machine Learning Research},
  Vol.~\bibinfo{volume}{119})}. \bibinfo{publisher}{{PMLR}},
  \bibinfo{pages}{9448--9457}.
\newblock


\bibitem[Trouillon et~al\mbox{.}(2016)]%
        {complex}
\bibfield{author}{\bibinfo{person}{Th{\'{e}}o Trouillon},
  \bibinfo{person}{Johannes Welbl}, \bibinfo{person}{Sebastian Riedel},
  \bibinfo{person}{{\'{E}}ric Gaussier}, {and} \bibinfo{person}{Guillaume
  Bouchard}.} \bibinfo{year}{2016}\natexlab{}.
\newblock \showarticletitle{Complex Embeddings for Simple Link Prediction}. In
  \bibinfo{booktitle}{\emph{{ICML}}} \emph{(\bibinfo{series}{{JMLR} Workshop
  and Conference Proceedings}, Vol.~\bibinfo{volume}{48})}.
  \bibinfo{publisher}{JMLR.org}, \bibinfo{pages}{2071--2080}.
\newblock


\bibitem[Vashishth et~al\mbox{.}(2020)]%
        {compgcn}
\bibfield{author}{\bibinfo{person}{Shikhar Vashishth}, \bibinfo{person}{Soumya
  Sanyal}, \bibinfo{person}{Vikram Nitin}, {and} \bibinfo{person}{Partha~P.
  Talukdar}.} \bibinfo{year}{2020}\natexlab{}.
\newblock \showarticletitle{Composition-based Multi-Relational Graph
  Convolutional Networks}. In \bibinfo{booktitle}{\emph{{ICLR}}}.
  \bibinfo{publisher}{OpenReview.net}.
\newblock


\bibitem[Vaswani et~al\mbox{.}(2017)]%
        {transformer}
\bibfield{author}{\bibinfo{person}{Ashish Vaswani}, \bibinfo{person}{Noam
  Shazeer}, \bibinfo{person}{Niki Parmar}, \bibinfo{person}{Jakob Uszkoreit},
  \bibinfo{person}{Llion Jones}, \bibinfo{person}{Aidan~N. Gomez},
  \bibinfo{person}{Lukasz Kaiser}, {and} \bibinfo{person}{Illia Polosukhin}.}
  \bibinfo{year}{2017}\natexlab{}.
\newblock \showarticletitle{Attention is All you Need}. In
  \bibinfo{booktitle}{\emph{{NIPS}}}. \bibinfo{pages}{5998--6008}.
\newblock


\bibitem[Veli{\v{c}}kovi{\'c} et~al\mbox{.}(2018)]%
        {gat}
\bibfield{author}{\bibinfo{person}{Petar Veli{\v{c}}kovi{\'c}},
  \bibinfo{person}{Guillem Cucurull}, \bibinfo{person}{Arantxa Casanova},
  \bibinfo{person}{Adriana Romero}, \bibinfo{person}{Pietro Lio}, {and}
  \bibinfo{person}{Yoshua Bengio}.} \bibinfo{year}{2018}\natexlab{}.
\newblock \showarticletitle{Graph attention networks}.
\newblock \bibinfo{journal}{\emph{ICLR}} (\bibinfo{year}{2018}).
\newblock


\bibitem[Vrandecic and Kr{\"{o}}tzsch(2014)]%
        {wikidata}
\bibfield{author}{\bibinfo{person}{Denny Vrandecic} {and}
  \bibinfo{person}{Markus Kr{\"{o}}tzsch}.} \bibinfo{year}{2014}\natexlab{}.
\newblock \showarticletitle{Wikidata: a free collaborative knowledgebase}.
\newblock \bibinfo{journal}{\emph{Commun. {ACM}}} \bibinfo{volume}{57},
  \bibinfo{number}{10} (\bibinfo{year}{2014}), \bibinfo{pages}{78--85}.
\newblock


\bibitem[Wang et~al\mbox{.}(2021c)]%
        {StAR}
\bibfield{author}{\bibinfo{person}{Bo Wang}, \bibinfo{person}{Tao Shen},
  \bibinfo{person}{Guodong Long}, \bibinfo{person}{Tianyi Zhou},
  \bibinfo{person}{Ying Wang}, {and} \bibinfo{person}{Yi Chang}.}
  \bibinfo{year}{2021}\natexlab{c}.
\newblock \showarticletitle{Structure-Augmented Text Representation Learning
  for Efficient Knowledge Graph Completion}. In
  \bibinfo{booktitle}{\emph{{WWW}}}. \bibinfo{publisher}{{ACM} / {IW3C2}},
  \bibinfo{pages}{1737--1748}.
\newblock


\bibitem[Wang et~al\mbox{.}(2018b)]%
        {RippleNet}
\bibfield{author}{\bibinfo{person}{Hongwei Wang}, \bibinfo{person}{Fuzheng
  Zhang}, \bibinfo{person}{Jialin Wang}, \bibinfo{person}{Miao Zhao},
  \bibinfo{person}{Wenjie Li}, \bibinfo{person}{Xing Xie}, {and}
  \bibinfo{person}{Minyi Guo}.} \bibinfo{year}{2018}\natexlab{b}.
\newblock \showarticletitle{RippleNet: Propagating User Preferences on the
  Knowledge Graph for Recommender Systems}. In
  \bibinfo{booktitle}{\emph{{CIKM}}}. \bibinfo{publisher}{{ACM}},
  \bibinfo{pages}{417--426}.
\newblock


\bibitem[Wang et~al\mbox{.}(2021b)]%
        {kepler}
\bibfield{author}{\bibinfo{person}{Xiaozhi Wang}, \bibinfo{person}{Tianyu Gao},
  \bibinfo{person}{Zhaocheng Zhu}, \bibinfo{person}{Zhengyan Zhang},
  \bibinfo{person}{Zhiyuan Liu}, \bibinfo{person}{Juanzi Li}, {and}
  \bibinfo{person}{Jian Tang}.} \bibinfo{year}{2021}\natexlab{b}.
\newblock \showarticletitle{{KEPLER:} {A} Unified Model for Knowledge Embedding
  and Pre-trained Language Representation}.
\newblock \bibinfo{journal}{\emph{Trans. Assoc. Comput. Linguistics}}
  \bibinfo{volume}{9} (\bibinfo{year}{2021}), \bibinfo{pages}{176--194}.
\newblock


\bibitem[Wang et~al\mbox{.}(2019a)]%
        {KGAT}
\bibfield{author}{\bibinfo{person}{Xiang Wang}, \bibinfo{person}{Xiangnan He},
  \bibinfo{person}{Yixin Cao}, \bibinfo{person}{Meng Liu}, {and}
  \bibinfo{person}{Tat{-}Seng Chua}.} \bibinfo{year}{2019}\natexlab{a}.
\newblock \showarticletitle{{KGAT:} Knowledge Graph Attention Network for
  Recommendation}. In \bibinfo{booktitle}{\emph{{KDD}}}.
  \bibinfo{publisher}{{ACM}}, \bibinfo{pages}{950--958}.
\newblock


\bibitem[Wang et~al\mbox{.}(2019b)]%
        {DBLP:conf/aaai/WangKMYTACFMMW19}
\bibfield{author}{\bibinfo{person}{Xiaoyan Wang}, \bibinfo{person}{Pavan
  Kapanipathi}, \bibinfo{person}{Ryan Musa}, \bibinfo{person}{Mo Yu},
  \bibinfo{person}{Kartik Talamadupula}, \bibinfo{person}{Ibrahim Abdelaziz},
  \bibinfo{person}{Maria Chang}, \bibinfo{person}{Achille Fokoue},
  \bibinfo{person}{Bassem Makni}, \bibinfo{person}{Nicholas Mattei}, {and}
  \bibinfo{person}{Michael Witbrock}.} \bibinfo{year}{2019}\natexlab{b}.
\newblock \showarticletitle{Improving Natural Language Inference Using External
  Knowledge in the Science Questions Domain}. In
  \bibinfo{booktitle}{\emph{{AAAI}}}. \bibinfo{publisher}{{AAAI} Press},
  \bibinfo{pages}{7208--7215}.
\newblock


\bibitem[Wang et~al\mbox{.}(2018a)]%
        {gcnz}
\bibfield{author}{\bibinfo{person}{Xiaolong Wang}, \bibinfo{person}{Yufei Ye},
  {and} \bibinfo{person}{Abhinav Gupta}.} \bibinfo{year}{2018}\natexlab{a}.
\newblock \showarticletitle{Zero-Shot Recognition via Semantic Embeddings and
  Knowledge Graphs}. In \bibinfo{booktitle}{\emph{{CVPR}}}.
  \bibinfo{publisher}{Computer Vision Foundation / {IEEE} Computer Society},
  \bibinfo{pages}{6857--6866}.
\newblock


\bibitem[Wang et~al\mbox{.}(2021a)]%
        {table1}
\bibfield{author}{\bibinfo{person}{Zhiruo Wang}, \bibinfo{person}{Haoyu Dong},
  \bibinfo{person}{Ran Jia}, \bibinfo{person}{Jia Li}, \bibinfo{person}{Zhiyi
  Fu}, \bibinfo{person}{Shi Han}, {and} \bibinfo{person}{Dongmei Zhang}.}
  \bibinfo{year}{2021}\natexlab{a}.
\newblock \showarticletitle{{TUTA:} Tree-based Transformers for Generally
  Structured Table Pre-training}. In \bibinfo{booktitle}{\emph{{KDD}}}.
  \bibinfo{publisher}{{ACM}}, \bibinfo{pages}{1780--1790}.
\newblock


\bibitem[Wang et~al\mbox{.}(2019c)]%
        {M-GNN}
\bibfield{author}{\bibinfo{person}{Zihan Wang}, \bibinfo{person}{Zhaochun Ren},
  \bibinfo{person}{Chunyu He}, \bibinfo{person}{Peng Zhang}, {and}
  \bibinfo{person}{Yue Hu}.} \bibinfo{year}{2019}\natexlab{c}.
\newblock \showarticletitle{Robust Embedding with Multi-Level Structures for
  Link Prediction}. In \bibinfo{booktitle}{\emph{{IJCAI}}}.
  \bibinfo{publisher}{ijcai.org}, \bibinfo{pages}{5240--5246}.
\newblock


\bibitem[Xian et~al\mbox{.}(2019)]%
        {awa}
\bibfield{author}{\bibinfo{person}{Yongqin Xian}, \bibinfo{person}{Christoph~H.
  Lampert}, \bibinfo{person}{Bernt Schiele}, {and} \bibinfo{person}{Zeynep
  Akata}.} \bibinfo{year}{2019}\natexlab{}.
\newblock \showarticletitle{Zero-Shot Learning - {A} Comprehensive Evaluation
  of the Good, the Bad and the Ugly}.
\newblock \bibinfo{journal}{\emph{{IEEE} Trans. Pattern Anal. Mach. Intell.}}
  \bibinfo{volume}{41}, \bibinfo{number}{9} (\bibinfo{year}{2019}),
  \bibinfo{pages}{2251--2265}.
\newblock


\bibitem[Xu et~al\mbox{.}(2022a)]%
        {inductive-ijcai}
\bibfield{author}{\bibinfo{person}{Xiaohan Xu}, \bibinfo{person}{Peng Zhang},
  \bibinfo{person}{Yongquan He}, \bibinfo{person}{Chengpeng Chao}, {and}
  \bibinfo{person}{Chaoyang Yan}.} \bibinfo{year}{2022}\natexlab{a}.
\newblock \showarticletitle{Subgraph Neighboring Relations Infomax for
  Inductive Link Prediction on Knowledge Graphs}. In
  \bibinfo{booktitle}{\emph{{IJCAI}}}. \bibinfo{publisher}{ijcai.org},
  \bibinfo{pages}{2341--2347}.
\newblock


\bibitem[Xu et~al\mbox{.}(2022b)]%
        {ENeSy}
\bibfield{author}{\bibinfo{person}{Zezhong Xu}, \bibinfo{person}{Wen Zhang},
  \bibinfo{person}{Peng Ye}, \bibinfo{person}{Hui Chen}, {and}
  \bibinfo{person}{Huajun Chen}.} \bibinfo{year}{2022}\natexlab{b}.
\newblock \showarticletitle{Neural-Symbolic Entangled Framework for Complex
  Query Answering}.
\newblock \bibinfo{journal}{\emph{CoRR}}  \bibinfo{volume}{abs/2209.08779}
  (\bibinfo{year}{2022}).
\newblock


\bibitem[Yao et~al\mbox{.}(2019)]%
        {kgbert}
\bibfield{author}{\bibinfo{person}{Liang Yao}, \bibinfo{person}{Chengsheng
  Mao}, {and} \bibinfo{person}{Yuan Luo}.} \bibinfo{year}{2019}\natexlab{}.
\newblock \showarticletitle{{KG-BERT:} {BERT} for Knowledge Graph Completion}.
\newblock \bibinfo{journal}{\emph{CoRR}}  \bibinfo{volume}{abs/1909.03193}
  (\bibinfo{year}{2019}).
\newblock


\bibitem[Yasunaga et~al\mbox{.}(2021)]%
        {qagnn}
\bibfield{author}{\bibinfo{person}{Michihiro Yasunaga}, \bibinfo{person}{Hongyu
  Ren}, \bibinfo{person}{Antoine Bosselut}, \bibinfo{person}{Percy Liang},
  {and} \bibinfo{person}{Jure Leskovec}.} \bibinfo{year}{2021}\natexlab{}.
\newblock \showarticletitle{{QA-GNN:} Reasoning with Language Models and
  Knowledge Graphs for Question Answering}. In
  \bibinfo{booktitle}{\emph{{NAACL-HLT}}}. \bibinfo{publisher}{Association for
  Computational Linguistics}, \bibinfo{pages}{535--546}.
\newblock


\bibitem[Ying et~al\mbox{.}(2021)]%
        {graphomer}
\bibfield{author}{\bibinfo{person}{Chengxuan Ying}, \bibinfo{person}{Tianle
  Cai}, \bibinfo{person}{Shengjie Luo}, \bibinfo{person}{Shuxin Zheng},
  \bibinfo{person}{Guolin Ke}, \bibinfo{person}{Di He},
  \bibinfo{person}{Yanming Shen}, {and} \bibinfo{person}{Tie{-}Yan Liu}.}
  \bibinfo{year}{2021}\natexlab{}.
\newblock \showarticletitle{Do Transformers Really Perform Badly for Graph
  Representation?}. In \bibinfo{booktitle}{\emph{NeurIPS}}.
  \bibinfo{pages}{28877--28888}.
\newblock


\bibitem[Zhang et~al\mbox{.}(2019)]%
        {itere}
\bibfield{author}{\bibinfo{person}{Wen Zhang}, \bibinfo{person}{Bibek Paudel},
  \bibinfo{person}{Liang Wang}, \bibinfo{person}{Jiaoyan Chen},
  \bibinfo{person}{Hai Zhu}, \bibinfo{person}{Wei Zhang},
  \bibinfo{person}{Abraham Bernstein}, {and} \bibinfo{person}{Huajun Chen}.}
  \bibinfo{year}{2019}\natexlab{}.
\newblock \showarticletitle{Iteratively Learning Embeddings and Rules for
  Knowledge Graph Reasoning}. In \bibinfo{booktitle}{\emph{{WWW}}}.
  \bibinfo{publisher}{{ACM}}, \bibinfo{pages}{2366--2377}.
\newblock


\bibitem[Zhang et~al\mbox{.}(2021)]%
        {pkgm}
\bibfield{author}{\bibinfo{person}{Wen Zhang}, \bibinfo{person}{Chi~Man Wong},
  \bibinfo{person}{Ganqiang Ye}, \bibinfo{person}{Bo Wen}, \bibinfo{person}{Wei
  Zhang}, {and} \bibinfo{person}{Huajun Chen}.}
  \bibinfo{year}{2021}\natexlab{}.
\newblock \showarticletitle{Billion-scale Pre-trained E-commerce Product
  Knowledge Graph Model}. In \bibinfo{booktitle}{\emph{{ICDE}}}.
  \bibinfo{publisher}{{IEEE}}, \bibinfo{pages}{2476--2487}.
\newblock


\bibitem[Zhang et~al\mbox{.}(2020)]%
        {hake}
\bibfield{author}{\bibinfo{person}{Zhanqiu Zhang}, \bibinfo{person}{Jianyu
  Cai}, \bibinfo{person}{Yongdong Zhang}, {and} \bibinfo{person}{Jie Wang}.}
  \bibinfo{year}{2020}\natexlab{}.
\newblock \showarticletitle{Learning Hierarchy-Aware Knowledge Graph Embeddings
  for Link Prediction}. In \bibinfo{booktitle}{\emph{{AAAI}}}.
  \bibinfo{publisher}{{AAAI} Press}, \bibinfo{pages}{3065--3072}.
\newblock


\end{thebibliography}

\end{document}